\documentclass[10pt,twocolumn,letterpaper]{article}

\usepackage{wacv}              

\usepackage{graphicx}
\usepackage{amsmath}
\usepackage{amssymb}
\usepackage{booktabs}
\usepackage{multirow}
\usepackage{caption}
\usepackage{float}
\usepackage{placeins}
\usepackage{color, colortbl}
\usepackage{stfloats}
\usepackage{enumitem}
\usepackage{tabularx}
\usepackage{xstring}
\usepackage{xspace}
\usepackage{url}
\usepackage{subcaption}
\usepackage{xcolor}

\usepackage{verbatim}
\usepackage{amsmath}
\usepackage{amssymb}
\usepackage{bm}
\usepackage{booktabs} 
\usepackage{multirow}
\usepackage{graphicx}
\usepackage{wrapfig}
\usepackage{array}   
\usepackage{wrapfig}
\usepackage{ulem}
\usepackage{CJKulem}
\usepackage{soul}
\usepackage{ulem} 
\usepackage{algorithm, algorithmicx, algpseudocode}

\usepackage{listings}
\usepackage{xcolor}
\usepackage[absolute,overlay]{textpos}
\usepackage[accsupp]{axessibility} 

\usepackage[pagebackref,breaklinks,colorlinks]{hyperref}

\usepackage[capitalize]{cleveref}
\crefname{section}{Sec.}{Secs.}
\Crefname{section}{Section}{Sections}
\Crefname{table}{Table}{Tables}
\crefname{table}{Tab.}{Tabs.}

\newcommand{\mrhtreplace}[2]{{\color{black} #2}}

\def\wacvPaperID{605} 
\def\confName{WACV}
\def\confYear{2025}

\begin{document}

\title{ACE: Anatomically Consistent Embeddings in Composition and Decomposition}

\author{Ziyu Zhou$^{*1,2}$ \quad Haozhe Luo$^{*2,3}$ \quad Mohammad Reza Hosseinzadeh Taher$^{2}$ \quad Jiaxuan Pang$^{2}$ \\ Xiaowei Ding$^{\dagger 1}$ \quad Michael Gotway$^{4}$ \quad Jianming Liang$^{\dagger 2}$\\
$^{1}$Shanghai Jiao Tong University \quad $^{2}$Arizona State University \quad $^{3}$University of Bern \quad $^{4}$Mayo Clinic\\
}
\maketitle

\begin{textblock*}{7cm}(1.8cm,25.5cm)  
    \noindent\rule[0.5ex]{5.2cm}{0.4pt} \\[0.5ex]  
    \footnotesize{$*$ Equal contribution. $\dagger$ Corresponding author.}
\end{textblock*}

\begin{abstract}

   Medical images acquired from standardized protocols show consistent macroscopic or microscopic anatomical structures, and these structures consist of composable/decomposable organs and tissues, but existing self-supervised learning (SSL) methods
   do not appreciate such composable/decomposable structure attributes inherent to medical images. To overcome this limitation, this paper introduces a novel SSL approach called ACE to learn {\bf \ul{a}}natomically {\bf \ul{c}}onsistent {\bf \ul{e}}mbedding via composition and decomposition with two key branches: (1) global consistency, capturing discriminative macro-structures via extracting global features; (2) local consistency, learning fine-grained anatomical details from composable/decomposable patch features via corresponding matrix matching. Experimental results across 6 datasets 2 backbones, evaluated in few-shot learning, fine-tuning, and property analysis, show ACE's superior robustness, transferability, and clinical potential. The innovations of our ACE lie in grid-wise image cropping, leveraging the intrinsic properties of compositionality and decompositionality of medical images, bridging the semantic gap from high-level pathologies to low-level tissue anomalies, and providing a new SSL method for medical imaging. All code and pretrained models are available at \href{https://github.com/jlianglab/ACE}{GitHub.com/JLiangLab/ACE}.
\end{abstract}




\begin{figure}[th!]
    \centering
    \setlength{\belowcaptionskip}{0.1cm}

    \includegraphics[width=8.2cm]{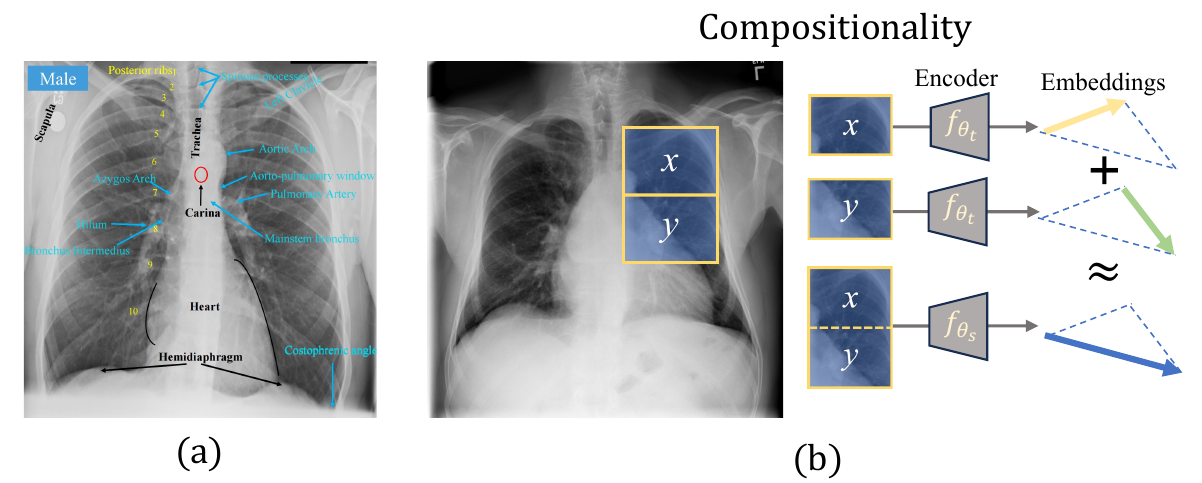}

    \caption{
(a) Chest X-rays contain various large (global) and small (local) anatomical patterns, including the right/left lung, heart, spinous processes, clavicle, mainstem bronchus, and the osseous structures of the thorax, which can be utilized for learning global and local embeddings in anatomy. (b) The hierarchical nature of anatomy (eg. The left lung has two lobes, the superior lobe $x$ and the inferior lobe $y$) calls for anatomical representation with compositionality where the embedding of the whole patch should be the sum of the embeddings of each part.
    }
    \label{fig:motivation}
    \vspace{-0.5cm}
\end{figure}

\section{Introduction}
\label{sec:intro}
\noindent Considering that expert labeling is costly and labor-intensive in medical imaging, self-supervised learning~\cite{jing2020self} has become a key approach towards annotation efficiency~\cite{zhou2017fine,taleb2021multimodal,tajbakhsh2021guest}. The recent SSL methods~\cite{caron2021emerging,grill2020bootstrap,chen2020simple,he2020momentum}, mainly designed for photographic images, do not fully exploit the inherent characteristics of \textbf{medical images}. Medical images (e.g., chest X-rays, fundus photography), unlike photographic images with target objects typically at the center on varying backgrounds, showcase consistent global (lung, heart) and local (clavicle, bronchus) anatomical structures as in Fig. \ref{fig:motivation}-a, resulting from standardized imaging protocols.
In addition, from an anatomical point of view, organs or tissues are decomposable, such as the left lung consisting of the superior lobe and the inferior lobe shown in Fig. \ref{fig:motivation}-b. To utilize these medical priors we hypothesize that \textit{simultaneously learning from global and local consistencies via composition and decomposition can equip the model to understand the anatomy, thereby offering strong transferability}.

To test this hypothesis, we have developed a framework that enables the capture of anatomical structures from unlabeled data, resulting in a powerful pretrained model. We call it \textbf{ACE} as it learns \textbf{\underline{a}}natomically \textbf{\underline{c}}onsistent \textbf{\underline{e}}mbedding via an innovative \ul{composition and decomposition} strategy, which was inspired by Hinton's idea paper~\cite{hinton2023represent} that people parse visual scenes and model the viewpoint-invariant spatial relationship between part and whole. Our framework consists of two novel components: learning global consistency and local consistency via composition and decomposition (detailed in Sec.~\ref{sec:method}).
Our ACE utilizes novel grid-wise image cropping, which differs from the existing random cropping strategy~\cite{islam2023self,caron2021emerging,bardes2022vicregl}, to provide truly precise patch matching.
Based on this cropping strategy, two randomly cropped views input to the student-teacher model in the global consistency branch are guaranteed to have overlaps, reducing the feature irrelevances~\cite{yuan2023densedino,zhang2022leverage}.
In the local consistency branch, %
we utilize the student-teacher model~\cite{grill2020bootstrap} to mimic the human understanding of part-whole relationships in images, where the embedding of a ``whole'' patch in one branch should always be consistent with the aggregated embedding of all ``part'' patches from the other branch, a process that we denote as composition and decomposition. We base this process on \textit{precise} patch matching, which differs significantly from the existing \textit{approximate} matching methods because they compute local consistency by narrowing the distance of semantically closest or spatially nearest features~\cite{bardes2022vicregl,xiao2021region,yuan2023densedino,yan2022sam,zhang2023precise}. ACE simultaneously optimizes a loss that integrates global and local consistencies to learn anatomically consistent embedding.

In this research, we mainly focus on chest X-rays (CXRs) and extensively evaluate ACE in (1) exploring the learned and emergent properties: ACE has been equipped with a set of unique properties by learning anatomies after pretraining (Sec. \ref{subsec:learned} and Sec. \ref{subsec:emergent}); (2) transferability to target tasks: ACE outperforms vision SSL methods designed for medical and photographic imaging and vision language SSL methods designed for medical imaging (Sec. \ref{subsec:downstream}); (3) generalization to other modality: ACE's adapting to fundus photography represents the universality to images acquired from standardized imaging protocols (Fig. \ref{fig:fundus}).
Our contributions lay in: 
\begin{itemize}[topsep=0pt, noitemsep]
    \item A new target for learning compositionality and decompositionality from unlabeled medical images, demonstrating that deep models can comprehend anatomical structures in human's way~\cite{hinton2023represent}.
    \item A novel exploration for pretrained backbone's properties including image retrieval, cross-patient anatomy correspondence, locality and symmetry of anatomical structures.
    \item A new SSL method with prominent transferability to various target tasks in medical image analysis.
\end{itemize}
  

\section{Related Work}
Self-supervised learning (SSL) methods share a common goal of learning meaningful representations without labeled data but differ in their primary learning focus: global features, local features, or structural patterns within images.

\noindent \textbf{Learning global features and local features.} A significant strand of SSL research concentrates on learning global features from images. These methods aim to capture the overall context of the image, ensuring consistency and alignment across different transformations of the input data. They can be grouped into two categories including contrastive learning~\cite{he2020momentum, chen2020simple, caron2020unsupervised, xiao2021region, tian2020contrastive} and non-contrastive learning methods~\cite{zbontar2021barlow, bardes2022variance, bardes2022variance, chen2021exploring, grill2020bootstrap, lee2021compressive}. 
This strand has progressively evolved to address challenges such as avoiding collapsing solutions~\cite{zhang2022does, chen2021exploring} and ensuring a balanced representation of global image features.  However, these methods typically excel in capturing the consistency of macro-structures, but they often fail to capture the complexity of local details,
compromising the accuracy needed for precise medical analysis. To address this, another research trajectory focuses on local feature learning whose objective is to distill fine-grained information by zooming in on specific parts of the image. This could involve detailed analysis of pixel-level features~\cite{xie2021propagate} or segmenting images into smaller, coherent regions~\cite{wang2021dense, bardes2022vicregl, xiao2021region, yun2022patch} to learn representations that honor the local semantic content of the visual input. 
Current approaches align feature vectors of patches that are semantically closest or spatially nearest neighbors~\cite{bardes2022vicregl, xiao2021region, yuan2023densedino}.
However, in these ways, the local patch embeddings may not be precisely paired, which may confuse the model. The previous work PEAC \cite{zhou2023learning} focused on local anatomical consistency through precise local matching, but only considered positive local patches, ignoring unpaired ones. In contrast, our ACE approach uses compositionality, decomposability, and grid-wise patch matching to align matched local embeddings and separate mismatched ones. Through learning our precise local consistency, our method helps models capture fine-grained details, improving disease segmentation and classification in medical images.

\noindent \textbf{Learning from structural patterns and anatomy.} Another avenue within SSL is the exploration of structural patterns and anatomy. Medical imaging holds consistent anatomical structures, particularly when using the same protocol, and naturally provides supervision signals for models to learn anatomical representations through self-supervision \cite{yan2022sam, zhou2019models}. Previous studies have focused on reconstructing anatomical patterns from transformed images \cite{zhou2019models}, understanding recurring anatomical patterns across different patients \cite{haghighi2020learning,haghighi2021transferable}, exploring spatial relationships within anatomy \cite{pang2022popar,pang2024asa,hosseinzadeh2023towards}, and enhancing these approaches with adversarial learning techniques \cite{guo2024stepwise,guo2022discriminative, haghighi2022dira,haghighi2024self,karani2020contrastive, fu2022anatomy, hu2022anatomy, jiang2023anatomical}. Although these methods focus on learning consistent representations of anatomy, they overlook the hierarchical relationships within anatomical structures. Additionally, the latest research Adam-v2~\cite{taher2024representing} exploits learning from composition and decomposition in a hierarchical way, but it remains limited to global patterns, overlooking the relationships in local patches.  
Unlike the existing methods, our ACE learns from anatomy by utilizing the consistency of global patterns and compositionality alongside the decomposability of local patterns, resulting in a hierarchical and integrative feature embedding structure.

\section{Our Method}
\label{sec:method}
\begin{figure*}[h!]
    \centering
    \setlength{\belowcaptionskip}{0.1cm}

    \includegraphics[width=16cm]{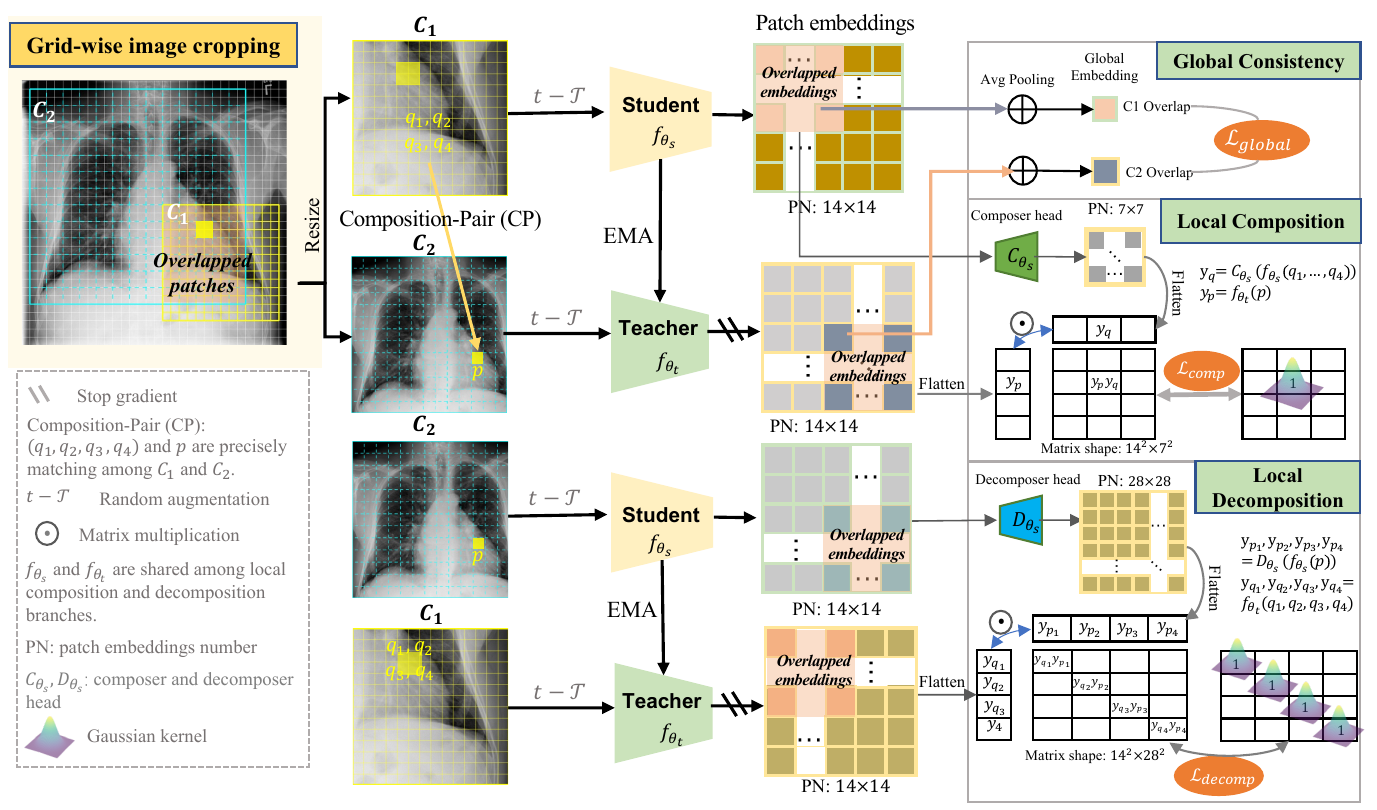}

    \caption{\mrhtreplace{}{ACE learns anatomically consistent embedding with two key branches: (1) global consistency and (2) local consistency via composition decomposition. Using our proposed grid-wise image cropping strategy (detailed in Sec. \ref{subsec:grid}), an input image is divided into a grid (see white grids), and two random crops, $C_1$ (yellow grids) and $C_2$ (green grids), are extracted. In the overlap region between $C_1$ and $C_2$, four patches in $C_1$ (denoted as $q_1, q_2, q_3, q_4$) correspond to one patch in $C_2$ (denoted as $p = {q_1, q_2, q_3, q_4}$).  The global consistency branch (detailed in Sec. \ref{subsec:global}) enforces consistency between the embeddings of the overlapping regions in $C_1$ and $C_2$ to learn coarse-grained semantic features. The local consistency branch (detailed in Sec. \ref{subsec:local_consistency}) enforces the model to learn fine-grained anatomical structure details via composition and decomposition. The local composition branch maximizes the similarity of paired patch embeddings and minimizes the similarity of unpaired ones to learn fine-grained anatomies in a part-to-whole manner. In a symmetrical process, the local decomposition branch enforces the model to learn fine-grained anatomies in a whole-to-part manner.}
    }
    \label{fig:architecture}
    \vspace{-0.3cm}
\end{figure*}

ACE, learning anatomies in global, local consistency via compositional and decompositional embeddings from unlabeled images, aims to bolster the development of self-supervised learning in medical imaging. The framework is shown in Fig. \ref{fig:architecture} which comprises two parts: (1) 
\textbf{global consistency} encourages the network to extract coarse-grained semantic features of different augmentations for the overlapped regions, (2) \textbf{local consistency} enforces the model to learn fine-grained local patterns via composition and decomposition. By integrating these components into a unified framework, ACE captures coarse to fine information in medical images, which provides powerful representations for various downstream tasks. In the following, we will introduce our methods from image pre-processing, each component to the joint training loss successively.

\subsection{Image pre-processing: grid-wise image cropping}
\label{subsec:grid}


\mrhtreplace{To obtain random image crops, we proposed a novel grid-wise image cropping strategy. As shown in Fig. \ref{fig:architecture}, an initial image is divided into grids $G$ (white grids) shaped $(32\times m)\times(32\times m)$, $m$ is the size of each grid. $C_2$ is randomly cropped from initial image with size of $(28\times m)\times(28\times m)$. And the upper-left point of $C_2$ is located at the node with coordinate $(x_2, y_2)$ of $G$. For $C_1$, the size is $(14\times m)\times (14\times m)$ and the beginning is the node of $C_2$ with upper-left coordinate $(x_1, y_1)$ in $G$. Using the coordinates $(x_1, y_1), (x_2,y_2)$ the indexes for overlap grids $idx_1=\{c_1,...,c_q\},idx_2=\{d_1,...,d_p\}$ of $C_1,C_2$ can be computed based on $G$, where $q=4p$ are the counts of overlap patches. For the overlapped region, each patch in $C_2$ precisely corresponds to 4 patches in $C_1$. In our experiments, $m$ is set to $32$ and the initial image size is $1024 \times 1024$.}{We introduce a grid-wise image cropping strategy to extract random image crops. First, the input image is divided into a grid $G$ of $32\times32$ non-overlapping patches (see white grids in Fig. \ref{fig:architecture}), with each patch having a size $m \times m$. Then, from the input image, we extract two random crops ($C_1$ and $C_2$ in Fig. \ref{fig:architecture}). $C_1$ comprises a $14\times14$ patch subset from $G$, and $C_2$ comprises a $28\times28$ patch subset from $G$. Based number of patches covered in each random crop of $C_1$ and $C_2$, in their overlap region, four patches in $C_1$, denoted as $q_1, q_2, q_3, q_4$, correspond to one patch in $C_2$, denoted as $p = \{q_1, q_2, q_3, q_4\}$. In the reverse case, in the overlap region between $C_1$ and $C_2$, each patch ($p$) in $C_2$ corresponds to four patches ($q_1, q_2, q_3, q_4$) in $C_1$. In our implementation, each patch has a size of $32\times32$ (i.e., $m = 32$) and the input image is $1024\times1024$ .}

\subsection{Learning global consistency}\label{subsec:global}
Global consistency motivates the network to extract consistent semantic features from various augmentations within the overlapping regions.
After grid-wise image cropping, the two crops are resized to the same shape $C_1, C_2 \in R^{C \times H_0 \times W_0}$, $C$ is the image channel and $H_0, W_0$ represent the height and width. The resized crops $C_1$ and $C_2$ are added different augmentations $x=\mathcal{T}_1(C_1),x^{'}=\mathcal{T}_2(C_2)$ where $\mathcal{T}_1,\mathcal{T}_2$ are two transformations.  Then the two crops are fed to the Student and Teacher models $f_{\theta_s}, f_{\theta_t}$ to get patch embeddings $y_{s}, y_t = f_{\theta_s}(x), f_{\theta_t}(x') \in R^{K \times N}$ respectively, where $K$ is the embedding dimension, $N=\frac{H_0 \times W_0}{m^2}$ and $(m,m)$ is the resolution of each patch. Then the average pooling operator $\oplus$ is added to the overlapped patches to generate global embeddings $y_{s\oplus}=Avg(y_s[O_1]), y_{t\oplus}=Avg(y_t[O_2]) \in R^{K}$, where $O_1$ and $O_2$ are overlapped area of $C_1,C_2$. The probability
distributions $P$ is obtained by normalizing the global embeddings with a softmax function: $P_s^{(i)}=\exp \left(y_{s\oplus}^{(i)} / \tau_s\right) / \sum_{k=1}^K \exp \left(y_{s\oplus}^{(k)} / \tau_s\right)$, 
 where $\tau_s > 0$ is a temperature parameter that controls the sharpness of the output distribution, and a similar formula holds for $P_t$ with temperature $\tau_t$. Given a fixed teacher network $f_{\theta_t}$, we learn to match these distributions by minimizing the cross-entropy loss:

\begin{equation}\label{eq:global}
    \mathcal{L}_{global} = \min _{\theta_s} CE(P_t, P_s)
\end{equation}

\noindent where $\theta_s$ the parameters of the student network, and $CE(a,b)=-alogb$.

\subsection{Learning local consistency}\label{subsec:local_consistency}
\noindent \textbf{Learning local consistency in composition.} The local composition encourages the model to learn fine-grained anatomies in a part-to-whole manner by encouraging consistency from the integration of sub-patches to a bigger patch. For the overlapped patches in the two crops, the Composition-Pair (e.g. $q_1, q_2, q_3, q_4$ and $p$ in Fig. \ref{fig:architecture}) are precisely matching, and the composition is defined:
\begin{equation}\label{eq:composition}
    y_q = C_{\theta_s}(f_{\theta_s}(q_1, q_2, q_3, q_4))
\end{equation}
\noindent where $y_q$ is the compositional representation of the 4 sub-patches, $C_{\theta_s}$ is composer head, and the bigger patch embedding $y_p = f_{\theta_t}(p)$, $f_{\theta_s}, f_{\theta_t}$ are student and teacher encoder.

We learn local consistency by maximizing the similarity between paired patch embeddings and minimizing it for unpaired ones. A CLIP~\cite{radford2021learning} like cross-correlation matrix is used to guide the model in learning this consistency. In detail, when $C_1$ is input to student and $C_2$ to teacher model, we get $N$ embeddings for each crop: $y_s, y_t \in R^{K \times N}$, and $y_s$ is input to composer head and get $C_{\theta_s}(y_s) \in R^{K \times N/4}$ to merge each $2\times 2$ embeddings into one embedding as Eq. \ref{eq:composition}. The cross-correlation matching matrix is defined as:
\begin{equation}
    M_{comp} = sigmoid(y_t^T \cdot C_{\theta_s}(y_s))
\end{equation}

\noindent where the matching matrix $M_{comp} \in R^{N\times N/4}$, $T$ is the transpose of a matrix, $(\cdot)$ is matrix multiplication, the $sigmoid$ function is added to restrict the values of the matrix to $(0,1)$. The value of the position $(i,j), M_{comp}^{(i,j)}=y_{t}[i]^{T} \cdot C_{\theta_s}(y_s)[j]$ represents the correlation between the two embeddings. Generally, the correlation weakens as the distance between their image patches increases. We use a Gaussian kernel $G(x,y)=\exp \left(-(x^2+y^2)/ (2 \sigma^2)\right)$ to smooth the matching matrix target, assigning a value of 1 to exact matches and decreasing values as the distance increases. We set $\sigma=1$ and kernel size $k=3$ for implementation. The target matrix $T_{comp} \in R^{N \times N/4}$ whose value in the position $(i,j)$:
\begin{footnotesize}
\begin{equation} \label{eq:target}
    T_{comp}^{(i,j)}=\left\{\begin{array}{l}
0, \text { if } |\Delta x^{(i,j)}|\text{ and } |\Delta y^{(i,j)}|>\frac{k-1}{2}\\
\exp \left(-\frac{|\Delta x^{(i,j)}|^2+|\Delta y^{(i,j)}|^2}{2}\right),  \text { others } 
\end{array}\right.
\end{equation}
 \end{footnotesize}

\noindent where $|\Delta x^{(i,j)}|,|\Delta y^{(i,j)}|$ are position distances between composed $C_1$ embedding $C_{\theta_s}(y_s)[j]$ and $C_2$ embedding $y_{t}[i]$ after overlapped area alignment. The local composition learning loss:
\begin{equation}
    \mathcal{L}_{comp} = \min _{\theta_s} \alpha CE(M_{comp}, T_{comp})
\end{equation}
\noindent the hyper-parameter $\alpha=0.9$ is used to balance the positive and negative samples.

\noindent \textbf{Learning local consistency in decomposition.} The local decomposition inspires the model learning consistency in a whole-to-part manner, which decomposes the patch embeddings into smaller sub-patches. As a symmetrical process of composition, decomposition learning lets inversely inputting: as shown in Fig. \ref{fig:architecture}, $C_2$ to student and $C_1$ to teacher model to get $y_s,y_t \in R^{K\times N}$. For a Composition-Pair, $p$ is decomposed into 4 sub-embeddings:
\begin{equation}\label{eq:decomposition}
    y_{p_{1}},y_{p_{2}},y_{p_{3}},y_{p_{4}} = D_{\theta_s}(f_{\theta_s}(p))
\end{equation}

\noindent the smaller patch embeddings from teacher $y_{q_{1}},y_{q_{2}},y_{q_{3}},y_{q_{4}}\\=f_{\theta_t}(q_1,q_2,q_3,q_4)$, where $f_{\theta_s}, f_{\theta_t}, D_{\theta_s}$ are student and teacher encoder, decomposer head. $y_s$ is input to the decomposer head to decompose each 1 to $2\times2$ embeddings. The cross-correlation matching matrix for decomposition $M_{decomp} = sigmoid(y_t^T \cdot D_{\theta_s}(y_s)) \in R^{N\times 4N}$. The decomposition matrix target $T_{comp} \in R^{N \times 4N}$ whose value in position $(i,j)$ is mathematical same with Eq. \ref{eq:target}. The local decomposition learning loss:
\begin{equation}
    \mathcal{L}_{decomp} = \min _{\theta_s} \alpha CE(M_{decomp}, T_{decomp})
\end{equation}
\noindent the hyper-parameter $\alpha=0.99$ is used to balance the positive and negative samples.

We illustrate ACE in Fig. \ref{fig:architecture} and propose a pseudo-code implementation of local consistency in Appendix Sec. A.1. The total loss is defined in Eq. \ref{eq:total loss}, where $\mathcal{L}_{global}$ is the global loss empowering the model to learn coarse-grained anatomical structure from global patch embeddings, $\mathcal{L}_{comp}, \mathcal{L}_{decomp}$ are two terms of local consistency loss equipping the model to learn precisely fine-grained local anatomical structures in composition and decomposition. $\lambda_1, \lambda_2, \lambda_3$ are coefficients to balance the weights of each loss term.
\begin{equation}\label{eq:total loss}
    \mathcal{L} = \lambda_1\mathcal{L}_{global}+\lambda_2 \mathcal{L}_{comp}+\lambda_3 \mathcal{L}_{decomp}
\end{equation}

\section{Implementation Details}
\noindent\textbf{Pretraining settings.} 
The composer and decomposer heads are 2-layer MLPs to integrate and expand the local embeddings.
The coefficient values of total loss are $\lambda_1=0.1, \lambda_2= \lambda_3 = 1$.
We pretrain our ACE on \mrhtreplace{unlabeled}{\textit{unlabeled}} ChestX-ray14~\cite{wang2017chestx} dataset with Swin-B~\cite{liu2021swin} and ViT-B~\cite{dosovitskiy2020image} backbones with $448^2$ images size training for 100 epochs. 
To compare with our proposed method, we take a variety of SSL methods developed for ResNet~\cite{he2016deep}, Vision Transformer~\cite{dosovitskiy2020image} and Swin-Transformer~\cite{liu2021swin} architectures. And these methods respectively leverage global information: DINO~\cite{caron2021emerging}, BYOL~\cite{grill2020bootstrap}; patch-level information: SelfPatch~\cite{yun2022patch}; and the structural patterns: Adam~\cite{hosseinzadeh2023towards}, POPAR~\cite{pang2022popar}, DropPos~\cite{wang2023droppos}. 
For equal comparison, we use the same experimental settings with ACE and pretrain these methods with ChestX-ray14 dataset. Besides, we also compare ACE with vision-language model KAD~\cite{zhang2023knowledge}, ChexZero~\cite{tiu2022expert} and DeViDe~\cite{luo2024devide} which are pretrained on chest X-ray images MIMIC-CXR dataset~\cite{johnson2019mimic}, and we directly load the pretrained models for downstream comparisons. More details are shown in Appendix Sec. A.3.

\noindent\textbf{Fine-tuning settings.}
We fine-tune pretrained models in supervised setting on downstream tasks including classification and segmentation. Classification performance is validated on 3 thoracic disease classification tasks ChestX-ray14~\cite{wang2017chestx}, Shenzhen CXR~\cite{jaeger2014two}, RSNA Pneumonia~\cite{rsna}.  For the segmentation task, we validate the dense prediction performance on JSRT~\cite{shiraishi2000development}, ChestX-Det~\cite{lian2021structure} and SIIM~\cite{siim-acr}. 
We transfer the pretrained models to each target task by fine-tuning the whole parameters.
The AUC (area under the ROC curve) metric is utilized to assess the performance of multi-label classification tasks 
on datasets such as ChestX-ray14 and Shenzhen CXR and for RSNA Pneumonia, we use accuracy as the evaluative measure.
For the target segmentation task, we use UperNet~\cite{xiao2018unified} as the training model and add an additional randomly initialized prediction head. The Dice is used to evaluate the segmentation performance. More Detailed settings including hyper-parameters are described in Appendix Sec. A.2, Sec. A.4.

\section{Results}

\mrhtreplace{To fully assess the properties of our framework, we conduct extensive experiments across quantitative metrics and qualitative indices, which could be divided into three categories: (1) learned properties shown in Sec. \ref{subsec:learned} that ACE has acquired from the training loss; (2) emergent properties shown in Sec. \ref{subsec:emergent} that ACE has not been trained in a specific manner but shows some capabilities; (3) downstream transferability to target tasks shown in Sec. \ref{subsec:downstream}.}{This section highlights the core results of our study, demonstrating the significance of our self-supervised learning framework, ACE. First, we present the properties that ACE was \textit{explicitly} trained to learn (Sec.~\ref{subsec:learned}). Next, we reveal the emergent properties of ACE that were \ul{not} part of its explicit training (Sec.~\ref{subsec:emergent}). Finally, we conduct extensive experiments to showcase ACE's generality and adaptability across various tasks (Sec.~\ref{subsec:downstream}), evaluated through two key aspects: (i) data efficiency, and (ii) fine-tuning settings.
}

\subsection{Learned properties} \label{subsec:learned}

\begin{figure}[h] 
    \centering
    \setlength{\belowcaptionskip}{0.1cm}
    \includegraphics[width=7cm]{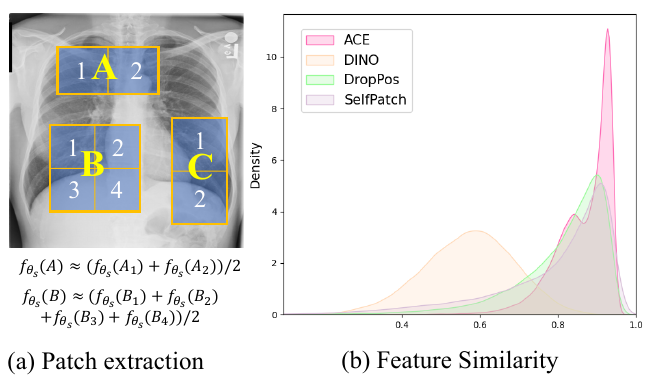}
    \caption{\mrhtreplace{ACE enhance feature composibility.}{ACE preserves the compositionality of anatomical structures in its learned embedding space. As seen, ACE's distribution is narrower and taller compared with DINO~\cite{caron2021emerging}, DropPos~\cite{wang2023droppos} and SelfPatch~\cite{yun2022patch}, with the mean similarity between embeddings of patches and their compositional parts closer to 1.}}
    \label{fig:comp}
    \vspace{-0.4cm}
\end{figure}

\noindent\textbf{(1) ACE enhances feature compositionality}\\
\mrhtreplace{To explore ACE's ability to comprehend compositionality of anatomical structures, we follow~\cite{hosseinzadeh2023towards} and set the experiment: randomly crop a region from an image and decompose it to 2 or 4 sub-patches shown in Fig. \ref{fig:comp_decomp} (a), then calculate cosine similarity between the embedding of the region and the average embedding of sub-patches, and report the values with Gaussian kernel density estimation (KDE) shown in Fig. \ref{fig:comp_decomp} (b). The distribution under our model $\text{ACE}_{s}$ is noticeably more peaked and shifted towards higher cosine similarity values, indicating a closer correspondence between the whole image embeddings and their compositional parts, underscoring the model's skillful integration of cohesive features while efficiently maintaining the compositional integrity of anatomical structures.}{
\noindent \textbf{Experimental Setup:} We investigate ACE's ability to preserve the compositionality of anatomical structures in its learned embedding space. Following~\cite{hosseinzadeh2023towards}, we randomly extract patches from \textit{test} images in the ChestX-ray14 dataset and further decompose each patch into 2 or 4 non-overlapping sub-patches. Each extracted patch and its sub-patches are then resized to a fixed dimension (\textit{i.e.}, $448 \times 448$) and their features are extracted using ACE's pretrained model as well as other baseline pretrained models. Subsequently, we calculate the \textit{cosine} similarity between the embedding of each patch and the average embedding of its sub-patches, and visualize the similarity distributions using Gaussian kernel density estimation (KDE).

\smallskip
\noindent \textbf{Results:} As shown in Fig. \ref{fig:comp}, ACE's distribution not only exhibits a narrower and taller shape compared with the baselines, but also has the mean similarity value between the embeddings of patches and their compositional parts (sub-patches) shifted towards 1. These observations demonstrate ACE's effective integration of cohesive features while maintaining the compositional integrity of anatomical structures, echoing its ability to preserve the compositionality of anatomical structures in its learned embeddings.
}

\noindent\textbf{(2) ACE enhances feature decompositionality.}

\begin{figure}[h]
    \centering
    \setlength{\belowcaptionskip}{0.1cm}
    \includegraphics[width=8cm]{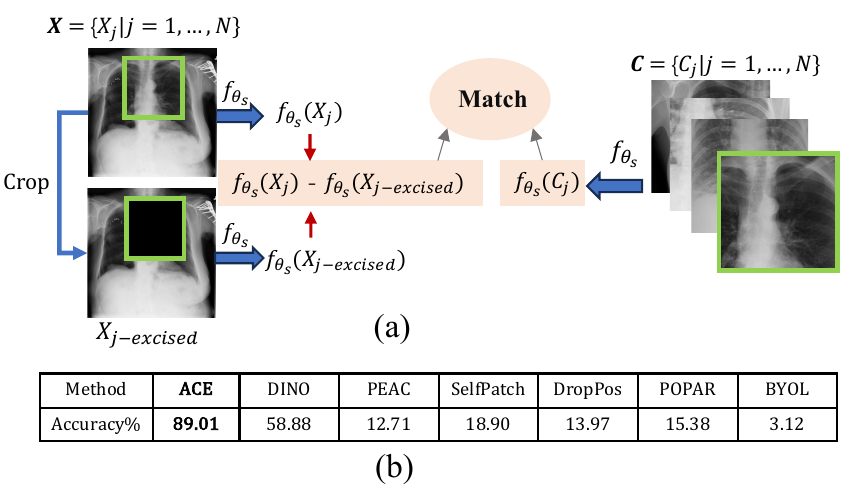}
    \caption{\mrhtreplace{ACE enhance feature decomposibility. (a) Examining the decompositionality process, (b) Matching results of ACE comparing with other baselines.\iffalse\mrhtreplace{}{(I cannot find $C_j$ in the Figure--I think the figure may need to be updated??)}\fi \iffalse\mrhtreplace{}{I think $X_j \in X$--above the image--- may be redundant??}\fi}{ACE preserves the decompositionality of anatomical structures in its learned embedding space. As seen, ACE outperforms SSL baselines by a large margin, achieving a high accuracy of 89.01\%, which is 30\% higher than the second-best baseline.}}
    \label{fig:decomp}
    \vspace{-0.2cm}
\end{figure}

\mrhtreplace{\textbf{Experimental Setup:} We explore the decomposibility property of our ACE learned for anatomical structures. Refer to Fig. \ref{fig:decomp}, local regions, 30-60\% of the original image size, are extracted from a batch of images, labeled as $C_j, j=1\dots N$, where $N$ is the batch size. \iffalse\mrhtreplace{}{(I cannot find $C_j$ in Fig. \ref{fig:decomp}--I think the figure may need to be updated??)}.\fi The model computes embeddings for the entire image \( X_j \), the image with a region removed \( X_{j-\text{excised}} \), and the excised region \( C_j \). The hypothesis for decomposition: $f_{\theta_s}(X_j) - f_{\theta_s}(X_{j-\text{excised}}) \approx f_{\theta_s}(C_j)$.The ChestX-ray14 test set is divided into batches, each containing 32 images. We calculate cosine similarity between $f_{\theta_s}(X_j) - f_{\theta_s}(X_{j-\text{excised}})$ and $f_{\theta_s}(C_j)$, finding out if the most similar value match $C_j$ correctly in each batch.}{
\noindent \textbf{Experimental Setup:} We examine ACE's ability to maintain the decompositionality of anatomical structures in its learned embedding space. To do so, we first divide the test set of the ChestX-ray14 dataset into batches, each containing 32 images. In each batch, we extract a random patch (from each image), labeled as $C_j$ in Fig. \ref{fig:decomp}--a, sized between 30-60\% of the original image. We then pass the original image, the image with the region removed (labeled as \( X_{j-\text{excised}} \)), and the extracted patches (excised region \( C_j \)) to the ACE's pretrained model and other baseline pretrained models to extract their features. Finally, we calculate the \textit{cosine} similarity between $f_{\theta_s}(X_j) - f_{\theta_s}(X_{j-\text{excised}})$ and $f_{\theta_s}(C_j)$, verifying if $f_{\theta_s}(X_j) - f_{\theta_s}(X_{j-\text{excised}}) \approx f_{\theta_s}(C_j)$.}

\smallskip
\mrhtreplace{\noindent \textbf{Results:} \iffalse\textcolor{green}{Do you have more baselines? If so, I would suggest presenting them in a ``table'' instead of comparing ACE with only DINO in the text??}\fi Our ACE model achieves accuracy at 89.01\%, surpassing other baselines shown in Fig. \ref{fig:decomp} (b): DINO 58.88\%, SelfPatch 18.90\%, DropPos 13.07\%, POPAR 15.38\% and BYOL 3.12\%. This notable increase in accuracy evidences the ACE model's superior capability in learning decomposable embeddings.}{
\noindent \textbf{Results:} As seen in Fig. \ref{fig:decomp}--b, ACE surpasses the SSL baselines by a remarkable margin. Notably, compared with DINO, PEAC, SelfPatch, DropPos, POPAR, and BYOL, which achieve accuracies of 58.88\%, 12.71\%, 18.90\%, 13.07\%, 15.38\%, and 3.12\% respectively, our ACE achieves a high accuracy of 89.01\%. This substantial difference in accuracy highlights ACE's superior ability to preserve the decompositionality of anatomical structures in its learned embeddings.
} 

\begin{figure}[h]
    \centering
    \setlength{\belowcaptionskip}{0.1cm}
    \includegraphics[width=7.5cm]{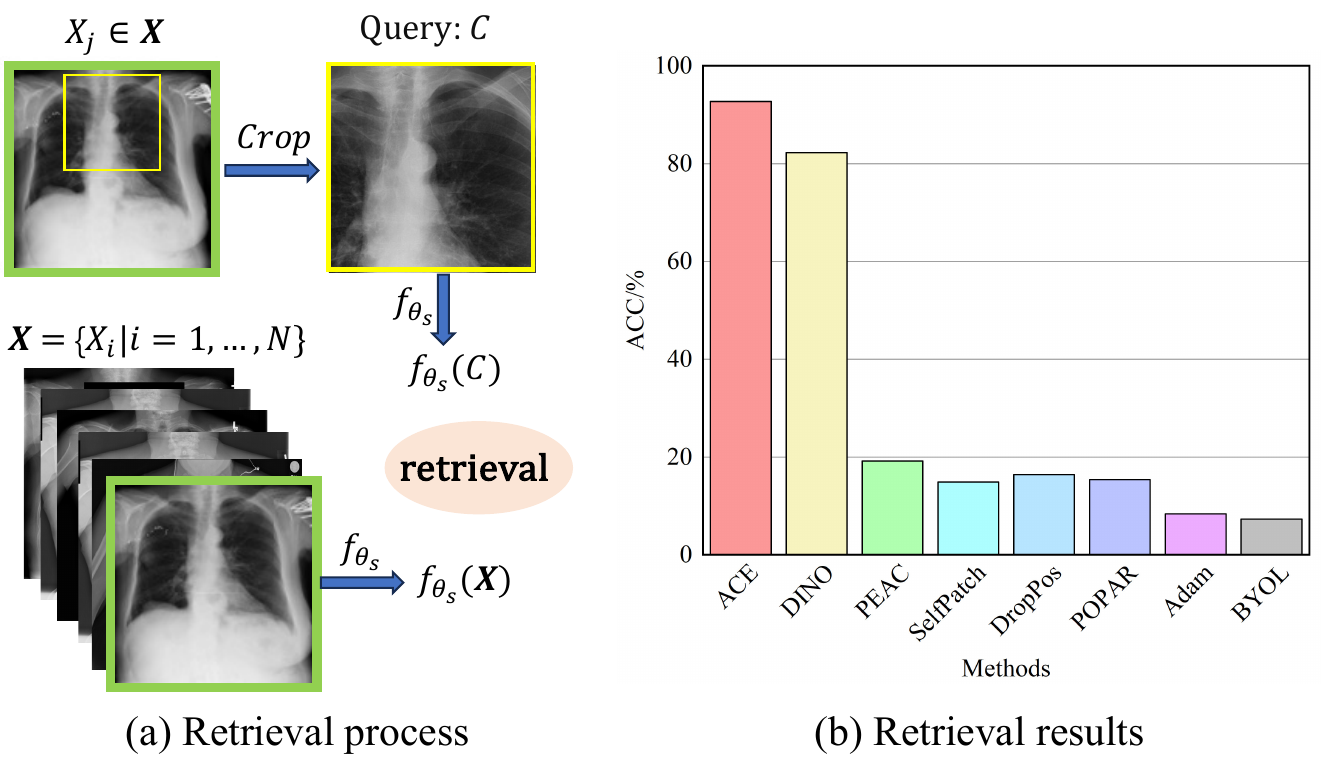}
    \caption{\mrhtreplace{(a) In the image retrieval process, $C$ is randomly cropped from an image and encoded as a query feature, which is used to retrieve the feature of the whole image in a batch. (b) ACE is compared with other baselines and shows superior performance in image retrieval.}{ACE captures semantics-rich features in its learned embedding space. As seen, ACE achieves higher retrieval accuracy compared with other SSL baselines.}}
    \label{fig:retrieval}
    \vspace{-0.4cm}
\end{figure}

\noindent\textbf{(3) ACE provides robust local feature-driven global image retrieval.}\\
\mrhtreplace{\textbf{Experimental Setup:} Consistent with decompositionality's image cropping setting, we take an image crop $C$ as the query image, the pretrained model $f_\theta_s$ (no fine-tuning) extracts the features of query $C$ and the batch of whole images $X$. The retrieval is then executed based on the cosine similarity scores between $f_{\theta_s}(C)$ and $f_{\theta_s}(X_i)$, where $i=1\dots N$ in a batch, selecting the image with the maximal similarity to $C$. The process is shown in Fig. \ref{fig:retrieval} (a).}{
\noindent \textbf{Experimental Setup:} We explore ACE's ability to capture semantics-rich features in its learned embedding space. To do so, we first divide the test set of the ChestX-ray 14 dataset into batches, each comprising 32 images. For each batch, we select a random image $X_j$ and extract a random patch from it to use as the \textit{query}, denoted as $C$ in~Fig. \ref{fig:retrieval}--a. Using ACE’s pretrained model and other baseline models, we extract features for the \textit{query} patch ($f_{\theta_s}(C)$) as well as for each whole image in the batch $ f_{\theta_s}(X_i) \mid X_i \in X $. We then compute the cosine similarity between the \textit{query} patch's embedding and the embeddings of the whole images in the batch. A retrieval is considered correct if the highest cosine similarity score corresponds to the \textit{query} patch and its associated whole image in the batch.}

\mrhtreplace{\textbf{Results:} Our approach achieves a significant lead in retrieval accuracy with a score of $94.37\%$ shown in Fig. \ref{fig:retrieval} (b). This performance is notably higher than competing methods, underscoring our model's potential applicability in enhancing diagnostic procedures such as using limited disease \jlnoe{space missing}{sample (disease prototypes)} for few-shot disease \jlnoe{same issue}{diagnosis (via matching)}. }{
\noindent \textbf{Results:} As seen in~Fig. \ref{fig:retrieval}--b, ACE achieves the highest retrieval accuracy ($94.37\%$) compared with other SSL baselines, demonstrating the semantic richness of ACE's learned representations. This result highlights ACE's clinical potential for accurately identifying and retrieving patients with similar pathological findings based on a query patch related to a specific disease.
}

\subsection{Emergent properties} \label{subsec:emergent}

We consider the following properties ``emergent'' as ACE is never trained with global and local consistencies across patients but such inter-image consistency has automatically emerged from training on intra-image consistency.

\noindent\textbf{(1) ACE provides distinctive anatomical embeddings}
\label{subsec:tsne}

\begin{figure}[h]
    \centering
    \setlength{\belowcaptionskip}{0.1cm}
    \includegraphics[width=7.5cm]{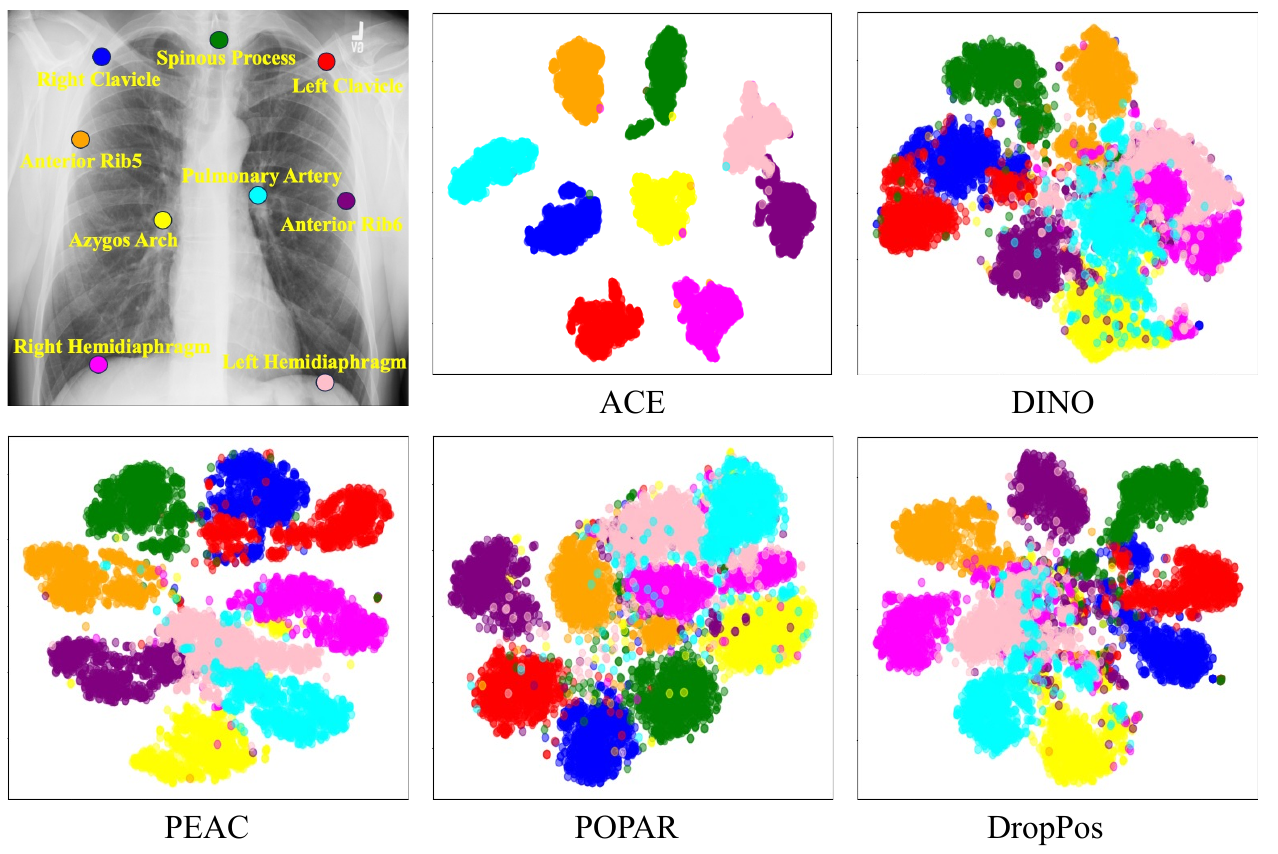}
    \caption{ 
\mrhtreplace{ACE provides distinctive anatomical embeddings which creates less overlaps cross different landmarks.}{ACE reflects locality of anatomical structures in its learned embedding space as an \textit{emergent} property. As seen, ACE, unlike the SSL baselines, distinguishes different anatomical structures in its embedding space while keeping identical anatomical structures across patients close to each other.}
    }
    \label{fig:tsne}
    \vspace{-0.4cm}
\end{figure}

\mrhtreplace{\textbf{Experimental Setup:} We explore ACE's capability in discriminating different anatomical structures by visualizing the embeddings. The dataset is the 1k images labeled 9 anatomical structures by human experts on \mrhtreplace{C1hestX-ray14}{ChestX-ray14} test set and the chosen landmarks are shown in Fig. \ref{fig:tsne}. We get crops of size $448^2$ around each landmark's location from the initial size of $1024^2$ images, and extract the local patch feature of each landmark using the pretrained model (with no fine-tuning). We then employ t-SNE~\cite{van2008visualizing} to visualize the embeddings of different landmarks.}{
\noindent \textbf{Experimental Setup:} We investigate ACE's ability to reflect the locality of anatomical structures in its learned embedding space. To do so, we compile a dataset of 1,000 images from the ChestX-ray14 dataset, each annotated by experts with 9 distinct anatomical landmarks (see~Fig. \ref{fig:tsne}). We extract $448^2$ patches around each landmark’s location from the $1024^2$ original images and then extract latent features for each landmark instance using ACE’s pretrained model and other baseline pretrained models (\textit{without} fine-tuning). These features are visualized using a t-SNE~\cite{van2008visualizing} plot.}

\smallskip
\mrhtreplace{\noindent \textbf{Results:} Our ACE shows superior discriminability in identifying anatomical landmarks compared to other baselines DINO~\cite{caron2021emerging}, PEAC~\cite{zhou2023learning}, POPAR~\cite{pang2022popar} and DropPos~\cite{wang2023droppos}. This sharp distinction is essential for accurately detecting anatomical landmarks in medical imagery, highlighting the refined embedding space of ACE for discrete local anatomical features. }{
\noindent \textbf{Results:} As seen in~Fig. \ref{fig:tsne}, the SSL baselines---DINO~\cite{caron2021emerging}, PEAC~\cite{zhou2023learning}, POPAR~\cite{pang2022popar} and DropPos~\cite{wang2023droppos}---struggle to generate distinct features for different landmarks, leading to ambiguous embedding spaces with mixed clusters. However, our ACE excels at distinguishing between various anatomical landmarks, resulting in well-separated clusters within its learned embedding space. This emphasizes ACE’s capability to develop a rich embedding space, where different anatomical structures are uniquely represented, and identical anatomical structures across patients have closely similar embeddings.}

\begin{figure}[h]
    \centering
    \setlength{\belowcaptionskip}{0.1cm}
    \includegraphics[width=8cm]{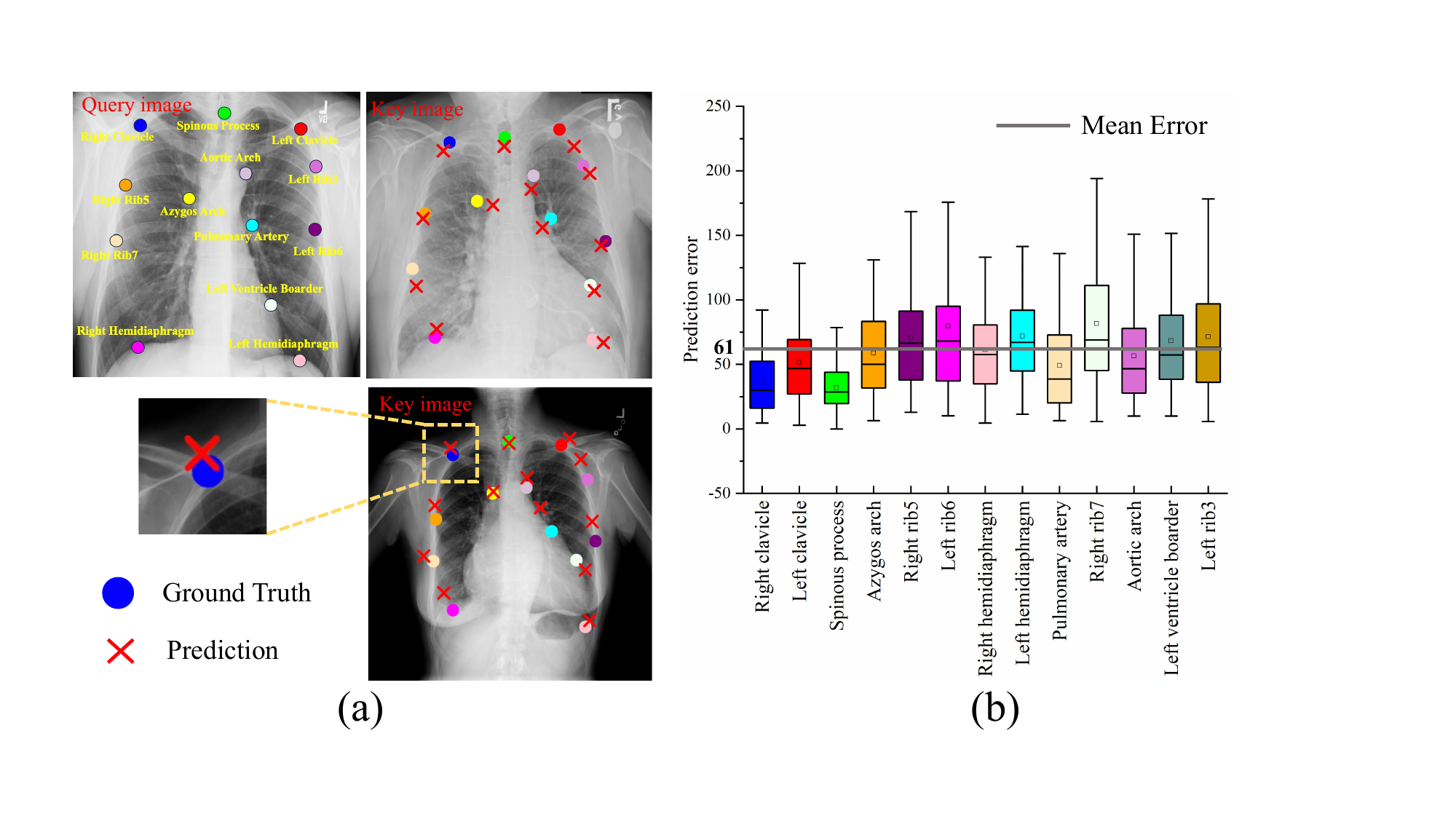}
    \caption{ 
ACE demonstrates its ability to accurately identify anatomical landmarks across different patients. (a) Zero-shot predictions and ground truth of selected query landmarks. (b) Quantitative analysis shows a low prediction error of 61 pixels in $1024^2$ images, highlighting the robustness of ACE model features in consistent cross-patient anatomical identification.
    }
    \label{fig:correspondence}
    \vspace{-0.4cm}
\end{figure}

\noindent\textbf{(2) ACE provides unsupervised cross-patient anatomy correspondence.} \\
\textbf{Experimental Setup}: To demonstrate the efficacy of our ACE in capturing a diverse range of anatomical structures, we utilize patch-level features to query the same anatomy across different patients in a zero-shot setting. In detail, we use the dataset mentioned in Sec. \ref{subsec:tsne} and choose $N_q=13$ landmarks labeled by human experts shown in Fig. \ref{fig:correspondence}-a. For a given query image, patches of size $448^2$ centered at each landmark point will be extracted from the initial size $1024^2$ image. These patches will be input to ACE's pretrained backbone (no fine-tuning) to get the query features of the centered landmarks $E_q = \{E_q^i\}_{i=1}^{N_q}$. Then for the rest key images, we extract $N_k$ patches by sliding a window of size $448^2$ with a stride of 8 (zero padding for the boundary patches), then input these patches to ACE backbone to get a dictionary of features for the key image $E_k = \{E_k^j\}_{j=1}^{N_k}$. Finally, for each query landmark feature in $E_q$, we find the closest feature in $E_k$ with $l_2$ distance and the position in the key image is the prediction corresponding landmark.

\noindent \textbf{Results:} We plot our predicting and ground truth landmarks in Fig. \ref{fig:correspondence}-a, and analyze the prediction errors with a box plot of each landmark shown in Fig. \ref{fig:correspondence}-b. From the results, the anatomical landmarks can be precisely detected using ACE encoded features and the average error of 13 landmarks is 61 pixels in 1k size of $1024^2$ images. Our findings indicate that the extracted features reliably represent specific anatomical regions and maintain consistency despite significant morphological variations. 

\subsection{Downstream transferability of ACE} \label{subsec:downstream}



\begin{figure}[h!]

    \centering
    \setlength{\belowcaptionskip}{0.1cm}

    \includegraphics[width=7.5cm]{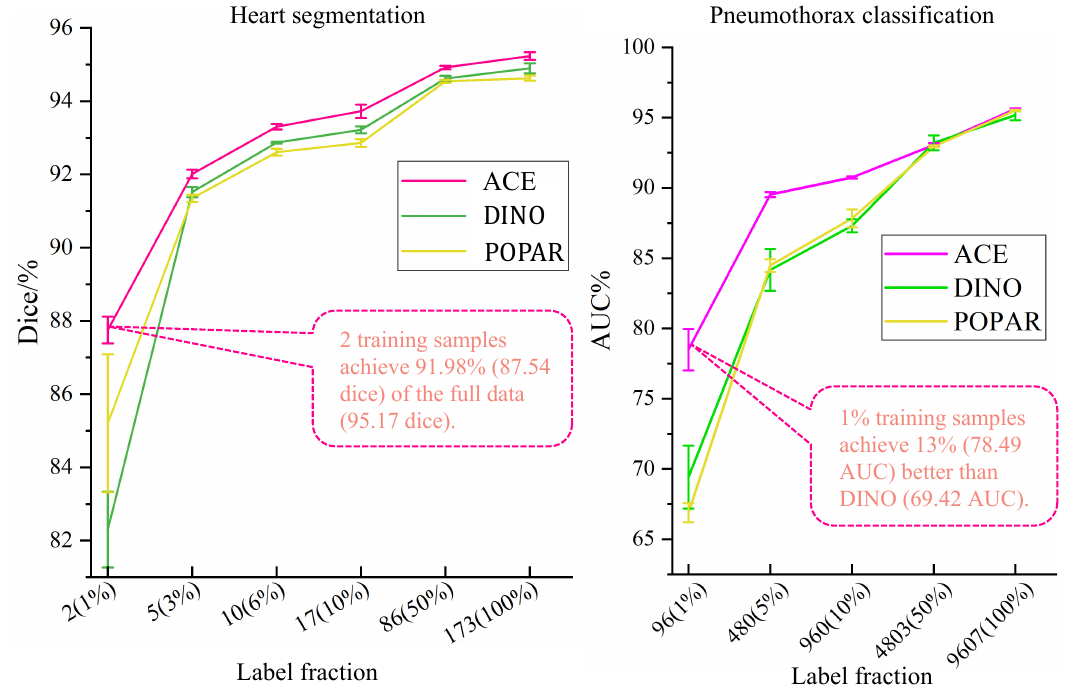}

    \caption{
    \mrhtreplace{ACE provides superior performance on data efficiency evaluation.}{
    ACE demonstrates superior performance in \textit{limited} data regimes. As seen in both heart segmentation and pneumothorax classification tasks, ACE surpasses SSL baselines (DINO and POPAR), particularly in few-shot transfer settings. 
    }}
    \label{fig:few-shot}
    \vspace{-0.5cm}

\end{figure}

\noindent\textbf{(1) Data efficiency evaluation}\\
\mrhtreplace{\noindent To investigate the robustness of representations learned by ACE, we fine-tune the pretrained model with limited labeled data and compare with other 2 SSL methods POPAR~\cite{pang2022popar} and DINO~\cite{caron2021emerging}. We conduct experiments on the heart segmentation dataset JSRT~\cite{shiraishi2000development} and pneumothorax classification dataset SIIM~\cite{siim-acr}, using a few shots of labeled data (2, 5 and 10) and smaller fraction data (10\% and 50\%). From the results, our ACE demonstrates increased gains with the reduction of training samples, and when fine-tuning only on \textbf{2} samples, ACE achieves 91.98\% performance of the full data.}{
\noindent \textbf{Experimental Setup:} We dissect robustness of ACE's representations in \textit{limited} data regimes. To do so, we compare pretrained ACE model with two SSL pretrained models, POPAR and DINO, by fine-tuning the models with a few labeled data (2, 5 and 10 shots) from the JSRT-Heart dataset~\cite{shiraishi2000development} and with limited labeled fractions (1\%, 10\% and 50\%) following~\cite{ma2023foundation} from the SIIM dataset ~\cite{siim-acr}.

\smallskip
\noindent \textbf{Results:} As seen in Fig. \ref{fig:few-shot}, ACE outperforms POPAR and DINO models in limited data regimes for both heart segmentation and pneumothorax classification tasks. Notably, in the heart segmentation task, ACE achieves over 91\% of its full training data performance using only 2 labeled samples. These results highlight ACE's annotation efficiency, particularly in target tasks where labeled data is scarce. 

}

\noindent\textbf{(2) Fine-tuning evaluation}\\
\mrhtreplace{\noindent To demonstrate the generalizability of ACE's representations, we fine-tune the pretrained model to a broad range of downstream tasks including classification, segmentation and key point detection. We compare the performances of classification and segmentation with models training from scratch and the other 9 self-supervised pretraining methods shown in Tab. \ref{tab:fine-tune}. From the results, we can get several observations: (\romannumeral1) our method surpasses the model training from scratch by a significant margin; (\romannumeral2) for the Swin-B backbone comparing with BYOL, DINO and POPAR, our ACE achieves the best performances among the 6 target classification, segmentation and the key point detection dataset; (\romannumeral3) for the ViT-B backbone, the performances of our ACE are outperforming or comparable among the ViT-B methods DINO, SelfPatch and DropPos; (\romannumeral4) Fig. \ref{fig:keypoint} shows accurate key point detection performance.}{
\noindent \textbf{Experimental Setup:} We investigate the generalizability of ACE's representations in full fine-tuning settings across a wide range of downstream tasks. To do so, we compare ACE with 9 SSL baselines with diverse objectives in 3 classification tasks and 4 segmentation tasks. Also, we include the performance of training from scratch as a lower bound.

\smallskip
\noindent \textbf{Results:} As seen in~Tab. \ref{tab:fine-tune}, our ACE model with the ViT-B backbone delivers competitive or superior performance compared with baselines with the same backbone, including DINO, SelfPatch, and DropPos. Moreover, ACE with the Swin-B backbone consistently outperforms BYOL, DINO, and POPAR, all of which also use the Swin-B backbone, across all downstream tasks. Comparing with methods adapted to medical imaging, including the vision SSL models POPAR, Adam and vision language SSL methods KAD, ChexZero, DeViDe, ACE can give the best or the second best performances. These results demonstrate the higher transferability and generalizability of ACE's representations across various tasks. 
}

 \begin{table*}[t]
        \centering
        \caption{\mrhtreplace{fine-tuning comparison with other self-supervised methods. The best methods are bolded, and the second best is underlined.}{ACE delivers generalizable representations for a broad spectrum of downstream tasks. As seen, ACE outperforms a diverse set of SSL baselines with different objectives and backbones across a range of classification and segmentation tasks in full fine-tuning settings. The best methods are bolded while the second best are underlined. In each task, we conducted the independent two-sample $t$-test between the best (bolded) vs. others. Highlighted boxes in blue indicate results which have no statistically significant difference at $p=0.05$ level.}}
        \label{tab:fine-tune}
        \resizebox{1.8\columnwidth}{!}{
        \begin{tabular}{cc|ccc|cccc}
        \toprule    
             Pretraining & \multirow{2}*{Backbone}& \multicolumn{3}{c|}{Classification} & \multicolumn{3}{c}{Segmentation}  \\
        \cline{3-9}
             Methods&  & ChestX-ray14  & ShenZhen & RSNA Pneumonia & JSRT clavicle & JSRT heart & ChestX-Det& SIIM  \\
        \midrule
            \rowcolor{gray!10}\multicolumn{9}{l}{\textit{Scratch}} \\
            \multirow{2}*{-}& Vit-B & $71.59 \pm 0.50$ & $83.18 \pm 0.55$& $63.17 \pm 0.33$ & $78.12 \pm 0.07$& $93.95\pm 0.09$& $60.04 \pm 0.41$ & $71.01 \pm 2.07$ \\
            & Swin-B & $79.28 \pm 0.07$& $84.61 \pm 0.81$ & $66.06 \pm 0.01$& $87.13 \pm 1.95$& $92.41 \pm 1.27$& $64.52 \pm 0.24$ & $73.69 \pm 0.66$  \\
        \midrule
            \rowcolor{gray!10}\multicolumn{9}{l}{\textit{Global features}} \\
            DINO~\cite{caron2021emerging} & ViT-B & $78.93 \pm 0.57$ & $94.79 \pm  0.47$&$71.08 \pm 0.19$ & $83.04 \pm 0.72$& $94.77\pm 0.39$ & $69.30 \pm 0.28$& $72.52 \pm 2.29$ \\
            BYOL~\cite{grill2020bootstrap} & Swin-B & $79.44 \pm 0.55$& $89.74 \pm 2.84$ & $73.43 \pm 0.20$& $90.14 \pm 0.42$& $94.80\pm 0.11$& $72.56 \pm 0.36$& $77.04 \pm 0.69$\\
            DINO~\cite{caron2021emerging} & Swin-B & $80.48 \pm 1.13$& $95.61 \pm 0.32$& $73.44 \pm 0.38$ & $91.74 \pm 0.06$& $94.90\pm 0.11$ & $72.25 \pm 1.50$& $78.89 \pm 0.50$\\ 
        \midrule
            \rowcolor{gray!10}\multicolumn{9}{l}{\textit{Local features}} \\
            SelfPatch~\cite{yun2022patch} &ViT-B & $79.27 \pm 0.81$& $95.42 \pm 0.84$&$70.94 \pm 1.03$ & $83.18 \pm 0.96$& $94.78\pm 0.34$ & $70.14 \pm 0.48$& $73.69 \pm 1.25$\\
        \midrule
            \rowcolor{gray!10}\multicolumn{9}{l}{\textit{Inherent structural patterns and anatomy}} \\
            Adam~\cite{hosseinzadeh2023towards} &ResNet50 & $81.81 \pm 0.14$ & $96.82 \pm 0.50$& $72.97 \pm 0.38$ & $\mathbf{93.30 \pm 0.07}$& $\cellcolor{cyan!25}\underline{95.33\pm 0.10}$ & $\underline{73.17 \pm 0.11}$ & $78.44 \pm 0.31$ \\
            DropPos~\cite{wang2023droppos} &ViT-B & $79.59 \pm 0.44$& $93.83 \pm 0.22$& $72.70 \pm 0.39$& $83.64 \pm 2.47$& $94.87\pm 0.37$& $69.69 \pm 2.03$& $70.52 \pm 2.55$\\
            POPAR~\cite{pang2022popar} &Swin-B & $81.84 \pm 0.02$ & $96.97 \pm 0.37$ &$73.78 \pm 0.07$& $91.77 \pm 0.08$& $94.63\pm 0.06$ &$73.01 \pm 0.39$& $78.65 \pm 0.18$ \\
            \multirow{2}*{$\text{PEAC}$} & ViT-B & $80.04 \pm 0.20$ & $96.69 \pm 0.30$ &$73.77 \pm 0.39$& $80.59 \pm 0.01$& $94.60\pm 0.08$ &$67.41 \pm 0.06$& $72.00 \pm 0.90$ \\
            & Swin-B & $\underline{82.78 \pm 0.21}$ & $97.39 \pm 0.19$ &$\cellcolor{cyan!25}\mathbf{74.39 \pm 0.66}$& $\underline{92.19 \pm 0.04}$& $95.21\pm 0.02$ &$73.01 \pm 0.16$& $\underline{78.97 \pm 0.26}$ \\
        \midrule
            \rowcolor{gray!10}\multicolumn{9}{l}{\textit{Vision language pretraining}} \\
            KAD~\cite{zhang2023knowledge} & ViT-B & $80.48 \pm 0.19$& $97.86 \pm 0.95$& $72.94 \pm 0.40$ & $81.67 \pm 0.06$& $95.07 \pm 0.12$& $69.9 \pm 0.30$& $72.51 \pm 1.78$\\ 
            ChexZero~\cite{tiu2022expert} & ViT-B & $78.67 \pm 0.07$& $96.34 \pm 0.13$& $72.70 \pm 0.21$ & $77.45\pm 0.20$ &$93.92 \pm 0.06$ & $68.41 \pm 0.37$& $76.41 \pm 0.76$\\
            DeViDe~\cite{luo2024devide} & ViT-B & $81.32 \pm 0.08$& $\cellcolor{cyan!25}\mathbf{98.03 \pm 0.24}$& $73.86 \pm 0.42$ & $81.73 \pm 0.13$& $94.85 \pm 0.03$ & $70.27 \pm 0.08$& $75.52 \pm 0.46$\\ 
        \midrule
            \rowcolor{gray!10}\multicolumn{9}{l}{\textit{Ours}} \\
            \mrhtreplace{$\text{ACE}_{v}$}{\multirow{2}*{$\text{ACE}$}} &ViT-B & $79.84 \pm 0.44$ &$95.73 \pm 0.12$ &$71.31 \pm 0.33$ & $83.66 \pm 0.56$ & $95.03 \pm 0.06$&$70.87 \pm 0.90$& $74.86 \pm 0.56$\\
            &Swin-B & $\mathbf{82.83\pm 0.08}$ & $\cellcolor{cyan!25}\underline{97.87 \pm 0.29}$ & $\cellcolor{cyan!25}\underline{74.31 \pm 0.16}$ & $91.91 \pm 0.07$& $\cellcolor{cyan!25}\mathbf{95.35 \pm 0.06}$&$\mathbf{74.37 \pm 0.42}$& $\mathbf{79.59 \pm 0.67}$  \\
        \bottomrule
        \end{tabular}}
        \vspace{-0.2cm} 
\end{table*}

\section{Ablation Study}

\begin{figure}[h!]

    \centering
    \vspace{0.05cm}
    \setlength{\belowcaptionskip}{0.1cm}

    \includegraphics[width=8cm]{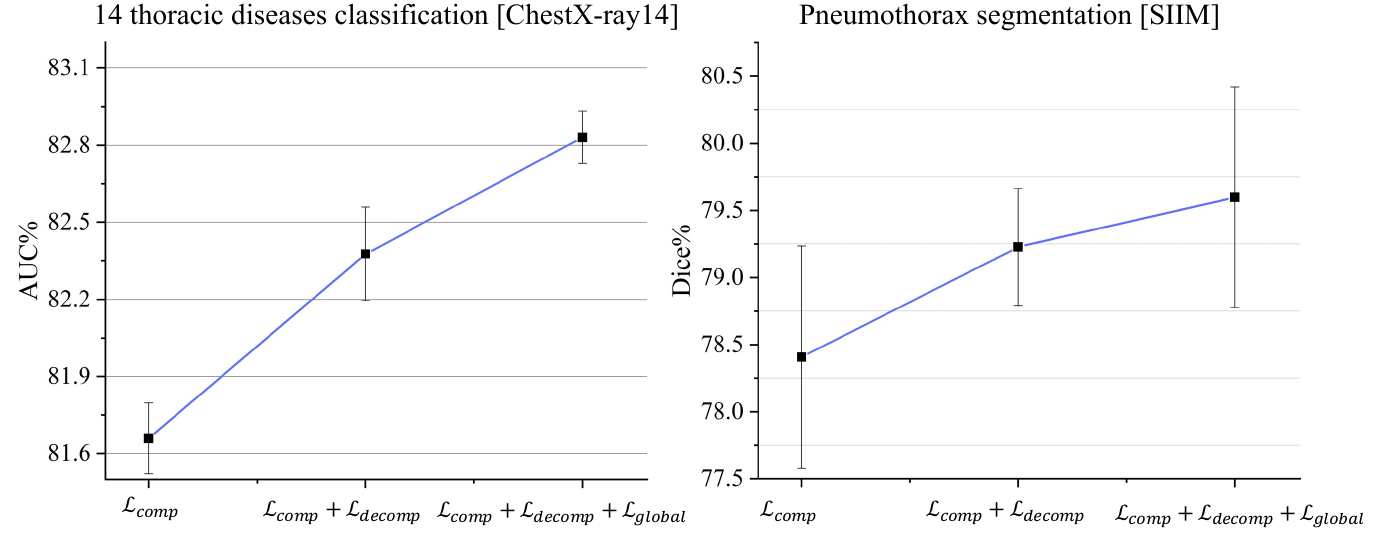}

    \caption{
Ablations on the learning objects. Adding the losses improves performance on classification and segmentation.
    }
    \label{fig:ablation}
    \vspace{-0.5cm}

\end{figure}

\noindent\textbf{Effectiveness of the learning Objects.}
We evaluate the impact of each learning component in ACE by progressively incorporating decompositionality and global loss, starting with compositionality. We fine-tune the models on two tasks: the ChestX-ray14 dataset for thoracic disease classification~\cite{wang2017chestx} and the SIIM dataset for pneumothorax segmentation~\cite{siim-acr}.  As shown in Fig. \ref{fig:ablation}, performance improves across both tasks with the addition of each loss component, highlighting the effectiveness of each.

\begin{figure}[h!]

    \centering
    \setlength{\belowcaptionskip}{0.1cm}

    \includegraphics[width=8.5cm]{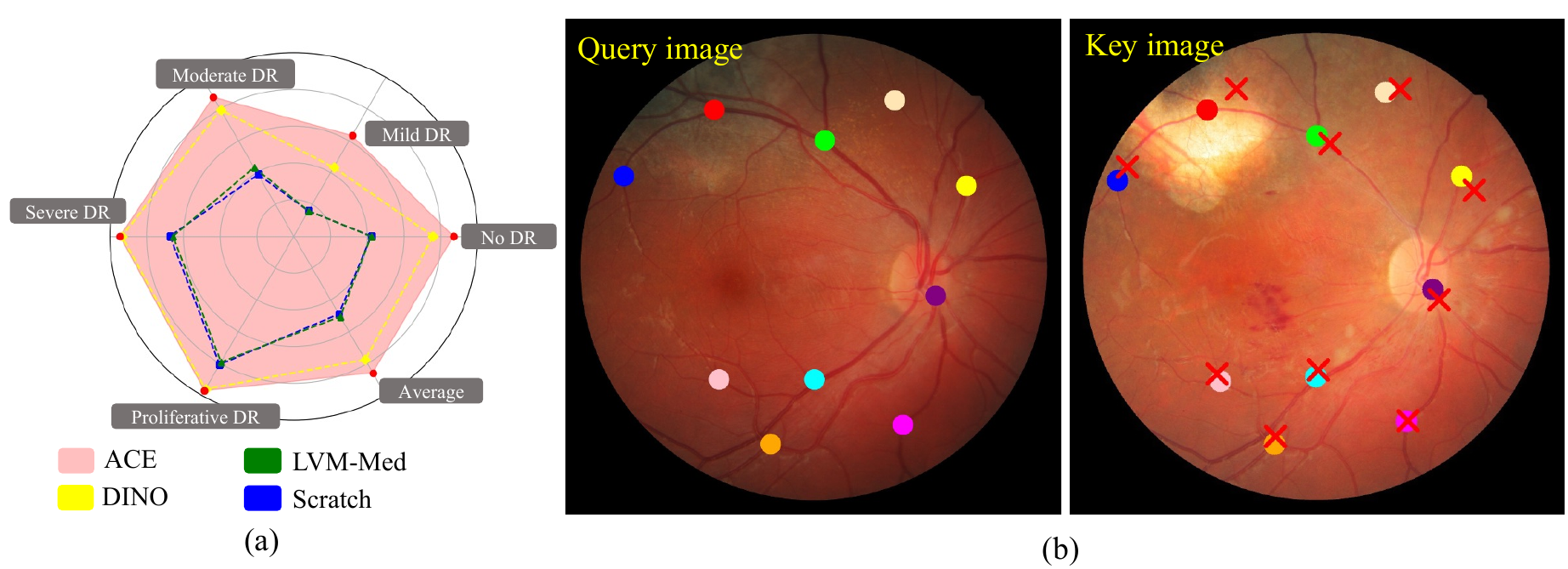}

    \caption{
(a) Comparison of proposed ACE with other baselines. AUC scores are shown on the EyePACS diabetic retinopathy classification dataset. (b) ACE's unsupervised landmark correspondence for paired left and rights fundus image on FIRE dataset. The colored rounds on query and key images are paired ground-truth landmarks, and the red crosses on key images are predicted ones.
    }
    \label{fig:fundus}
    \vspace{-0.5cm}

\end{figure}

\noindent\textbf{Generalizability of ACE.} Our framework can be seamlessly extended to other imaging modalities. To demonstrate this, we pretrain ACE on unlabeled fundus images EyePACS~\cite{cuadros2009eyepacs} and fine-tune it on diabetic retinopathy classification dataset EyePACS. As seen in Fig. \ref{fig:fundus}-a, ACE exhibits superior performance compared with SOTA SSL method DINO~\cite{yuan2023densedino}, large-scale pretraining method LVM-Med~\cite{mh2024lvm} and training from scratch. Besides, we conduct unsupervised anatomy correspondence based on ACE's pretrained backbone without fine-tuning on fundus image registration dataset FIRE~\cite{hernandez2017fire}. As shown in Fig. \ref{fig:fundus}-b, the key-points corresponding with the query image can be precisely located in the key image.

\section{Conclusion}
We introduce ACE, a novel SSL method aimed at visual representation learning via composition and decomposition for anatomical structures in medical images. Our method relies on learning global consistency and local consistency by reliable global representation alignment and correspondence matrix matching. ACE has been rigorously tested through comprehensive experiments in various tasks, demonstrating its emergent properties and effective transferability, showing significant promise for advancing and explainable AI applications in medical image analysis.

\section{Acknowledgement}
This research has been supported in part by ASU and Mayo Clinic through a Seed Grant and an Innovation Grant, and in part by the NIH under Award Number R01HL128785. The content is solely the responsibility of the authors and does not necessarily represent the official views of the NIH. This work has utilized the GPUs provided in part by the ASU Research Computing and in part by the Bridges-2 at Pittsburgh Supercomputing Center through allocation BCS190015 and the Anvil at Purdue University through allocation MED220025 from the Advanced Cyberinfrastructure Coordination Ecosystem: Services \& Support (ACCESS) program, which is supported by National Science Foundation grants \#2138259, \#2138286, \#2138307, \#2137603, and \#2138296. The content of this paper is covered by patents pending. 

\bigskip

\appendix

\section*{Supplementary Materials}
\section{Implementation Details}

\subsection{Pseudo code implementation}

We illustrate the method of ACE in the main paper and propose a pseudo-code implementation of local consistency in the Algorithm. \ref{code:local consis}.

\begin{algorithm}[h]
   \caption{local consistency Pytorch pseudo-code.}
   \label{code:local consis}
    \definecolor{codeblue}{rgb}{0.25,0.5,0.5}
    \lstset{
      basicstyle=\fontsize{7.2pt}{7.2pt}\ttfamily\bfseries,
      commentstyle=\fontsize{7.2pt}{7.2pt}\color{codeblue},
      keywordstyle=\fontsize{7.2pt}{7.2pt},
    }
\begin{lstlisting}[language=python]
# fs, ft: student and teacher encoders
# Cs, Ds: composer and decomposer head
# O1, O2: overlap area mask (0/1) of crop C1 and C2
# kernel: gaussian kernel to smooth the matrix target
ft.params = fs.params
ft.requires_grad = False
for C1, C2, O1, O2 in loader: # load a minibatch
    C1, C2 = augment(C1), augment(C2) # random views

    s1, s2 = fs(C1), fs(C2) # student output
    t1, t2 = ft(C1), ft(C2) # teacher output

    # compute composition loss
    s1 = Cs(s1) # input composer head
    O1 = maxpool(O1) # pool size 2x2
    loss_comp = ComputeLoss(O1, O2, s1, t2)

    # compute decomposition loss
    s2 = Ds(s2) # input composer head
    O2 = interpolate(O2) # upsample the overlap mask
    loss_decomp = ComputeLoss(O1, O2, s1, t2)

    loss = loss_comp/2 + loss_decomp/2
    loss.backward() # back-propagate

    # student, teacher updates
    update(fs) # Adam
    ft.params = m*gt.params + (1-m)*gs.params

def ComputeLoss(O1, O2, s, t):
    # compute matching matrix
    M = torch.mul(t.flattern(), s.flattern())
    
    # compute matrix target
    T = torch.zeros(len(t.flattern()),len(s.flattern()))  
    idx1 = torch.nonzero(O1)
    idx2 = torch.nonzero(O2)

    # apply gaussian weights on the target coordinates
    T[idx2, idx1] = kernel 
        
    return - (T * log(M)).sum(dim=1).mean()

\end{lstlisting}
\end{algorithm}

\subsection{Pretraining and testing datasets}

We evaluate our ACE on chest X-rays and fundus photography, pretraining on ChestX-ray14~\cite{wang2017chestx} and EyePACS~\cite{cuadros2009eyepacs} datasets respectively. The pretrained ACE models are validated on target tasks including the following datasets: 
\begin{itemize}[topsep=0pt, noitemsep]
    \item {\bf ChestX-ray14}~\cite{wang2017chestx}, which contains 112K frontal-view X-ray images of 30805 unique patients with the text-mined fourteen disease image labels (where each image can have multi-labels). We use the official training set 86K (90\% for training and 10\% for validation) and testing set 25K. The downstream models are trained to predict 14 pathologies in a multi-label classification setting and the mean AUC score is utilized to evaluate the classification performance. In addition to image-level labeling, the datasets provides bounding box annotations for 880 images in test set. Of this set of images, bounding box annotations are available for 8 out of 14 thorax diseases. After finetuning, we use the bounding box annotations in test set to assess the accuracy of pathology localization in a weakly-supervised setting. Besides, we compile a dataset of 1,000 images from test set, each annotated by experts with distinct anatomical landmarks. We use these labeled landmarks for anatomical embeddings analysis (see main paper Sec. 5.2-1 and Sec. \ref{subsec:symmetry}), unsupervised key-point correspondence (see main paper Sec. 5.2-2), key-point detection (see Sec. \ref{subsec:keypoint})
    \item {\bf NIH Shenzhen CXR}~\cite{jaeger2014two}, which contains 326 normal and 336 Tuberculosis (TB) frontal-view chest X-ray images. We split 70\% of the dataset for training, 10\% for validation and 20\% for testing which are the same with ~\cite{ma2022benchmarking};
    \item {\bf RSNA Pneumonia}~\cite{rsna}, which consists of 26.7K frontal view chest X-ray images and each image is labeled with a distinct diagnosis, such as Normal, Lung Opacity and Not Normal (other diseases). 80\% of the images are used to train, 10\% to valid and 10\% to test. 
    \item \textbf{JSRT}~\cite{shiraishi2000development}, which is an organ segmentation dataset including 247 frontal view chest X-ray images. All of them are in 2048$\times$2048 resolution with 12-bit gray-scale levels. The heart and clavicle segmentation masks are utilized for this dataset. We split 173 images for training, 25 for validation and 49 for testing.
     \item \textbf{ChestX-Det}~\cite{lian2021structure}, which is a disease segmentation dataset and an improved version of ChestX-Det10~\cite{liu2020chestx}. This dataset contains 3,578 images with instance-level annotations for 13 common thoracic pathology categories, sourced from the NIH ChestX-ray14 dataset. Annotations were provided by three board-certified radiologists, and the dataset includes additional segmentation annotations. We consolidated all the diseases into one region and the goal of segmenting this dataset is to distinguish between diseased and non-diseased areas for each image. There are official split for training and testing sets and we split 10\% images from training set for validation.
    \item \textbf{SIIM-ACR}~\cite{siim-acr}, a dataset resulting from a collaboration between SIIM, ACR, STR, and MD.ai, contains 12,089 chest X-ray images. It is the largest public pneumothorax segmentation dataset to date, comprising 3,576 pneumothorax images and 9,420 non-pneumothorax images, all of which are available in 1024x1024 pixel resolution. We randomly divided the dataset into training (80\%), validation (10\%) and testing (10\%). The segmentation performance is measured by the mean Dice which average the dice of pneumothorax non-pneumothorax images.
    \item \textbf{EyePACS}~\cite{cuadros2009eyepacs}, a  diabetic retinopathy (DR) classification dataset for identifying signs of diabetic retinopathy in eye images. The clinician has rated the presence of diabetic retinopathy in each image on a scale of 0 to 4, 0 for no DR, 1 for mild DR, 2 for moderate DR, 3 for severe DR and 4 for proliferative DR. There are 53,576 unlabeled images and 35,126 with labels. We randomly split the labeled images into training (80\%), validation (10\%) and testing (10\%) set for downstream evaluation, and we merge the training, validation and unlabeled sets for pretraining.
    \item \textbf{FIRE}~\cite{hernandez2017fire}, the dataset comprises 134 pairs of images obtained from 39 patients, with each pair annotated with specific corresponding keypoints. In our target task, for each pair of images, one image is designated as the query image, and the task is to identify the corresponding anatomical structures in the key image. Additionally, we simultaneously visualize the predicted and ground truth keypoints in the key image.
\end{itemize}
\subsection{Pretraining settings}


We have trained two ACE models with Swin-B backbone using unlabeled images from ChestX-ray14 and EyePACS for the adaptation on chest X-ray and fundus imaging. Moreover, to generalize to other architecture we have trained ACE on ViT-B backbone on ChestX-ray14.
Our ACE learning paradigm is similar to knowledge distillation~\cite{hinton2015distilling}, where a student network learns to match a teacher network's output. 
The weights of the student model $\theta_s$ are updated by back-propagation and the gradients of teacher model are stopped whose weights $\theta_t$ are updated using EMA (exponential moving average) from student. The update rule is $\theta_t \leftarrow \lambda \theta_t+(1-\lambda) \theta_s$, where $\lambda$ follows a cosine schedule from 0.996 to 1 during training.

The composer and decomposer heads are 2-layer MLPs  to integrate and expand the local embeddings. In detail, the output of student or teacher encoder are patch embeddings with shape $14\times 14 \times 1024$. Before the composer head, each $2\times 2\times 1024$ adjacent embeddings are concatenated and the patch embeddings are reshaped to $7\times 7 \times 4096$, then they are input to a 2-layer MLP with input dimension 4096 and output dimension 1024 to get shape of $7\times 7 \times 1024$ embeddings. Symmetrically, in the decomposer head, the $14\times 14 \times 1024$ patch embeddings are input to a 2-layer MLP with input dimension 1024 and output dimension 4096 to expand the embeddings to $14\times 14 \times 4096$, then each embedding is chunked into $2\times 2\times 1024$ and the output embeddings will be $28\times 28 \times 1024$. 

During the pretraining phase, we utilize a batch size of 8 images per GPU and train for a total of 100 epochs with 4 V100 (32G). The optimizer is AdamW and the initial learning rate is set to 5e-4 with a linear warm-up over the first 10 epochs. The weight decay starts at 0.04 and reaches 0.4 by the end of training, following a cosine schedule. The drop path rate is set to 0.1. Gradient clipping is applied with a maximum norm of 0.8 to ensure stable training dynamics. 

\subsection{Finetuning settings}
For the target classification tasks, we concatenate a randomly initialized linear layer to the output of the classification (CLS) token of ViT-B pretrained models. For Swin-B pretrained models, we add an average pooling to the last-layer feature maps, then feed the feature to the randomly initialized linear layer. For the target segmentation task, we use UperNet~\cite{xiao2018unified} as the training model. We concatenate pretrained weights and randomly initialized prediction head for segmenting. 
Following ~\cite{liu2021swin}, we employ the AdamW optimizer in conjunction with a cosine learning rate scheduler. We incorporate a linear warm-up phase spanning 20 epochs, within a total training duration of 150 epochs. The base learning rate is set at 0.0001. Each experiment is conducted using four V100 32 GPUs, with a batch size of 32 per GPU. For segmentation tasks, we retain the same setup and extend the training period to 500 epochs.

\section{Additional Results}

\subsection{Emergent property: ACE understand anatomical symmetry.}\label{subsec:symmetry}

\begin{figure}[h]
    \centering
    \setlength{\belowcaptionskip}{0.1cm}
    \includegraphics[width=8cm]{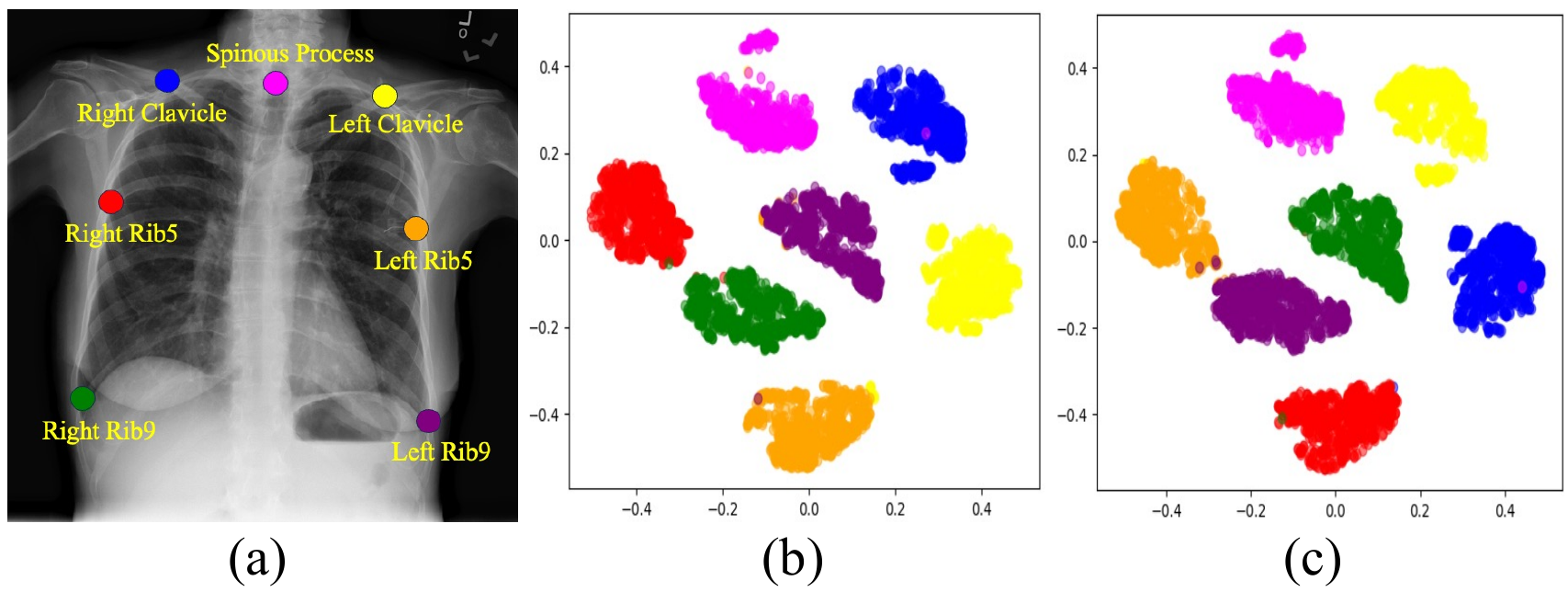}
    \caption{\mrhtreplace{Validation of ACE understanding the anatomical symmetry. For the (a) chosen anatomical landmarks, we compare (b) t-SNE visualization and (c) left and right flipping patches visualization. The anatomies show similar features with their mirrored symmetrical ones, e.g., Right Rib5 (red cluster in (b)) and left and right flipped Left Rib5 (orange cluster in (c)) hold similar features. }{ACE reflects the symmetry of anatomical structures in its learned embedding space as an \textit{emergent} property. As seen, ACE provides mirrored embeddings for mirrored anatomical structures (e.g., the right and left clavicles, and the right and left rib 5.).}}
    \label{fig:symmetry}
\end{figure}

\noindent \textbf{Experimental Setup:} We examine ACE's ability to capture the symmetry of anatomical structures in its learned embedding space. To do so, we consider $N=7$ anatomical landmarks, including three pairs of mirrored structures and one structure located in the center of the chest, as shown in~Fig. \ref{fig:symmetry}--a. We extract size of $448^2$ patches ($C = \{C_i\}_{i=1}^N$) around each landmark’s location from the original images, and then use ACE's pretrained model to extract latent features for each landmark and its corresponding left and right flipped version ($\Tilde{C}=\mathcal{T}(C)$). The extracted features of $C$ and $\Tilde{C}$ are visualized via t-SNE plots in~Fig. \ref{fig:symmetry}--b and \ref{fig:symmetry}--c, respectively.

\smallskip
\noindent \textbf{Results:} As seen in~Fig.\ref{fig:symmetry}--b and Fig.\ref{fig:symmetry}--c, ACE captures the symmetry of anatomical structures within its learned embedding space. For example, the right and left clavicles, which are visually symmetrical, are represented similarly in the embedding space.  As seen, the blue cluster in~Fig.\ref{fig:symmetry}--b, corresponding to the right clavicle, closely matches the yellow cluster in~Fig. \ref{fig:symmetry}--c, which represents the flipped left clavicle. A similar pattern is observed for other pairs, such as the left rib 5 and its flipped version, represented by the orange and red clusters in~Fig.\ref{fig:symmetry}--b and Fig.\ref{fig:symmetry}--c, respectively. These observations demonstrate that ACE effectively captures the symmetry of anatomical structures in its learned embedding space as an \textit{emergent} property.

\subsection{Fine-tuning evaluation: key point detection}\label{subsec:keypoint}

\begin{figure}[h!]

    \centering
    \setlength{\belowcaptionskip}{0.1cm}

    \includegraphics[width=8.5cm]{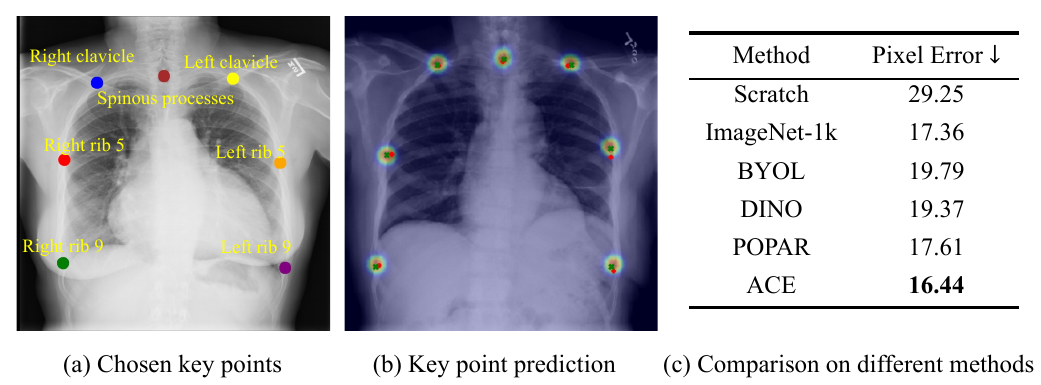}

    \caption{
ACE demonstrates its ability to boost downstream key point detection tasks. (a) 7 key points are chosen for fine-tuning; (b) the inference detection image where the \textcolor{red}{red} points are prediction while \textcolor{green}{green} points are the ground truth situating at the center of the heatmap; (c) comparison between ACE and other pretrained baselines and the lower pixel error the better detection performance.
    }
    \label{fig:keypoint}

\end{figure}

\noindent\textbf{Experimental Setup:} we investigate the generalizability of ACE's pretrained model via fine-tuning the landmark detection task. To do so, we use the dataset annotated by experts with distinct anatomical landmarks (mentioned in main paper Sec. 5.2), and we choose 7 key points as shown in Fig. \ref{fig:keypoint}-a. We load the pretrained weights of ACE and other baselines including ImageNet-1K, BYOL, DINO and POPAR. The fine-tuning architecture is UperNet which is the same with segmentation, while the training target is the specific points of interest. Following~\cite{sun2019deep}, we optimize the detection process based on the heatmap method, that is, we add a $11\times 11$ Gaussian kernel $\exp \left(-\frac{x^2+y^2}{2 \sigma^2}\right)$ to smooth each ground truth landmark where the peak is 1 and the values decrease as the distance increase. The learning target is visualized in Fig. \ref{fig:keypoint}-b where the green points are the center of the heatmaps. The error between prediction and ground truth points is used as the evaluation metric.

\noindent\textbf{Results:} As seen in Fig. \ref{fig:keypoint}-c, initializing with our ACE's weights can get the lowest pixel error 16.44 while the image size is $448\times448$, better than initialized with other baselines, ImageNet pretrained weights and training from scratch. From the results, ACE's representations can give some priors about the anatomical structure which boosts to distinguish the key points. 

\begin{figure*}[h!]
    \centering
    \setlength{\belowcaptionskip}{0.1cm}
    \includegraphics[width=17.5cm]{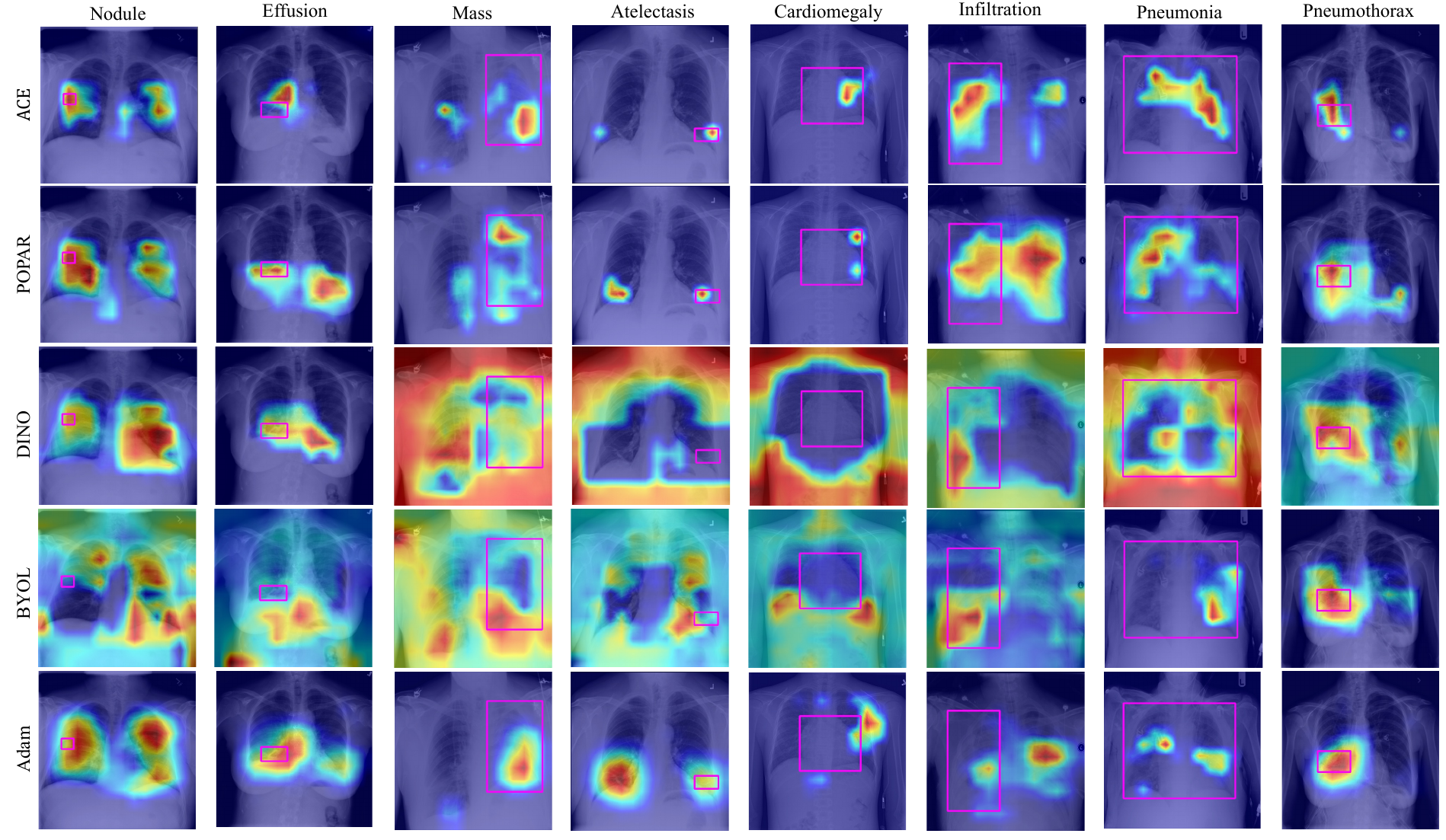}
    
\caption{Visualization of Grad-CAM heatmaps. For each column, we provide the heatmap examples for 8 thorax diseases that hold bounding boxes in official labeling. The first row shows the results of our method ACE while the rest of the rows represent the localization of POPAR, DINO, BYOL and Adam. The pink boxes represent the localized ground truth.}
\label{supple:gradcam}
\end{figure*}

\subsection{Weakly supervised localization}

\noindent\textbf{Experimental setup:} To compare with other pretraining methods POPAR~\cite{pang2022popar}, DINO~\cite{caron2021emerging}, BYOL~\cite{grill2020bootstrap} and Adam~\cite{hosseinzadeh2023towards}, we initialize downstream model with these pretrained weights using only image-level disease label on ChestX-ray14 dataset. After fine-tuning, the models are used for inference on 787 cases annotated with bounding boxes for eight thorax diseases: Atelectasis, Cardiomegaly, Effusion, Infiltrate, Mass, Nodule, Pneumonia, and Pneumothorax. We use Grad-CAM~\cite{selvaraju2017grad} heatmap to approximate the localization of a specific thorax disease predicted by the trained model. The baseline Adam is fintuned on ResNet50 and other methods are based on Swin-B. 

\noindent\textbf{Results:} Fig. \ref{supple:gradcam} shows the visualization of heatmaps generated by ACE, POPAR, DINO, BYOL and Adam for 8 thorax pathologies in ChestX-ray14 dataset. From the results, the localization of our method surpasses the learning global feature methods DINO and BYOL and learning inherent structure pattern method POPAR and Adam. For analyzing the Grad-CAM heatmaps, our method shows more precise and compact localization with small shifts, while the learning global feature methods DINO and BYOL often completely can not localize the diseases. Surprisingly, our model can also localize some small pathologies like nodules and atelectasis, which demonstrate the positive impact of the combination of learning global and local anatomies.

{\small
\bibliographystyle{ieee_fullname}
\normalem
\bibliography{egbib}
}

\end{document}